\documentclass[journal]{IEEEtran}
\usepackage[nocompress]{cite}  
\usepackage{amsmath,amssymb,amsfonts}
\usepackage{algorithmic}
\usepackage{graphicx}
\usepackage{textcomp}
\usepackage{booktabs} 
\usepackage{array} 
\usepackage{algorithm}
\usepackage{amsthm}
\theoremstyle{plain}
\newtheorem{theorem}{Theorem}[section]
\newtheorem{proposition}[theorem]{Proposition}

\theoremstyle{definition}

\newtheorem{assumption}[theorem]{Assumption}
\theoremstyle{remark}
\newtheorem{remark}[theorem]{Remark}
\usepackage{hyperref}
\usepackage[capitalize,noabbrev]{cleveref}
\usepackage{subcaption}
\usepackage{float}

\begin{document}

\title{ Understanding the Role of Rehearsal Scale in Continual Learning under Varying Model Capacities }

\author{

JinLi He, Liang Bai, Xian Yang

\IEEEcompsocitemizethanks{ 
\IEEEcompsocthanksitem J. He and L. Bai are with Institute of Intelligent Information Processing, Shanxi University, Taiyuan, 030006, China (Corresponding author: Liang Bai)\protect\\
Email: hejinli@sxu.edu.cn, bailiang@sxu.edu.cn
\IEEEcompsocthanksitem X. Yang is with Alliance Manchester Business School, The University of Manchester, Manchester, M13 9PL, UK. Email: xian.yang@manchester.ac.uk   }

}

\maketitle

\begin{abstract}

    Rehearsal is one of the key techniques for mitigating catastrophic forgetting and has been widely adopted in continual learning algorithms due to its simplicity and practicality. However, the theoretical understanding of how rehearsal scale influences learning dynamics remains limited. To address this gap, we formulate rehearsal-based continual learning as a multidimensional effectiveness–driven iterative optimization problem, providing a unified characterization across diverse performance metrics. Within this framework, we derive a closed-form analysis of adaptability, memorability, and generalization from the perspective of rehearsal scale. Our results uncover several intriguing and counterintuitive findings. First, rehearsal can impair model's adaptability, in sharp contrast to its traditionally recognized benefits. Second, increasing the rehearsal scale does not necessarily improve memory retention. When tasks are similar and noise levels are low, the memory error exhibits a diminishing lower bound. Finally, we validate these insights through numerical simulations and extended analyses on deep neural networks across multiple real-world datasets, revealing statistical patterns of rehearsal mechanisms in continual learning.    
   
\end{abstract}


\section{Introduction}
\label{section1_introduction}

    \IEEEPARstart{I}{ntelligent} systems need to acquire, update, and accumulate knowledge throughout their lifecycle to adapt to the dynamically changing real world, a capability known as continual learning \cite{2continual1robotics1995,2continual2aaai1986}. Typically, continual learning machines are challenged by catastrophic forgetting \cite{3catastrophic1psychology1989,3catastrophic2arXiv2013,3catastrophic3iclr2020}, where  performance on previous tasks degrades dramatically due to parameter updates when learning new tasks. As new knowledge replaces previous knowledge, the model's adaptation performance improves while memorability diminishes \cite{4tradeoff1neurosciences2005,4tradeoff2zhengzehua2cvpr2022,4tradeoff3cvpr2023}. Earlier efforts have attempted to address this problem by preserving previously learned knowledge \cite{replay1GEM2017,5forget1structure4cvpr2021,5forget3zhengzehua5cvpr2023}. However, recent work has focused more on facilitating the adaptability of new knowledge and the generalizability of models \cite{ktwoplayergame2021,6generate1cvpr2022,kforgetgenerate2023}. These efforts have deepened the understanding of continual learning: an ideal continual learning learner should strike an effective and delicate balance between retaining previously learned knowledge and acquiring new knowledge, while also being sufficiently generalizable to accommodate differences in unseen data distributions.

    In biological systems, hippocampal replay \cite{7hippocampa1neuron2009,7hippocampa4science2025} has been proposed as a system-level mechanism that consolidates memories and improves the generalization by reactivating previously experienced scenes. Although biological and artificial systems differ significantly, they exhibit intriguing parallels: both consolidate knowledge and accelerate learning from past experiences \cite{8rephippocampa1neurps2017,replay2nature2020,8rephippocampa3nature2025}. Similar to biological systems, rehearsal mechanism performs better in continual learning to resist catastrophic forgetting \cite{9improvereplay1ecv2018,9improvereplay2replay3cvpr2022,9improvereplay3replay4icml2023,replay5neurps2024,replay2nature2020}. Despite advances in the empirical performance of rehearsal-based methods, the theoretical understanding of how the rehearsal mechanism impacts continual learning, even in simple models, is not yet fully understood: When does a continual learning model benefit from rehearsal? Will the rehearsal scale itself adversely impact performance?

    \graphicspath{{picture/} }
    \begin{figure*}[t]
        \centering    
        \includegraphics[width=1\linewidth]{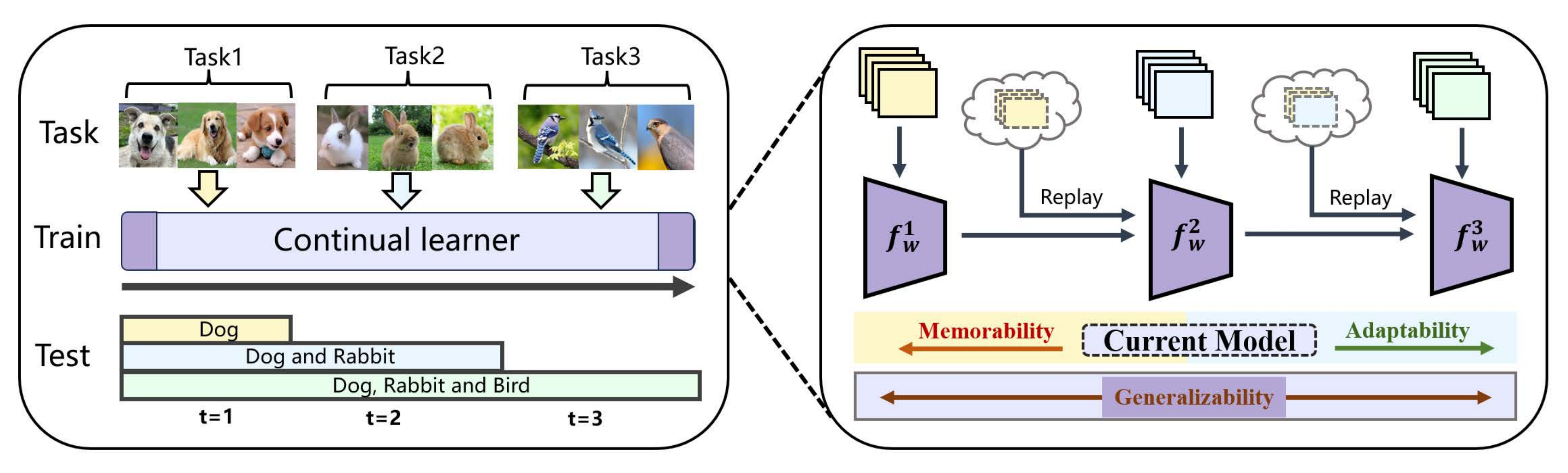}   
        \caption{An ideal continual learning system should strike a delicate balance among the adaptation of newly acquired knowledge, the memorization of previously learned knowledge, and the generalization of unseen data distributions across diverse scenarios.}    
        \label{图片1_表述图}    
    \end{figure*}

    In this paper, we address the aforementioned questions by theoretically elucidating the limitations of rehearsal scale. We establish a theoretical characterization of rehearsal-based continual learning from three complementary perspectives, highlighting both its strengths and inherent drawbacks. The main contributions are summarized as follows.

    \begin{itemize}
    \item We derive closed-form expressions characterizing theoretical error along three evaluation dimensions, revealing how rehearsal scale influences performance. 
    \item We demonstrate that increasing the rehearsal scale does not necessarily lead to better performance: rehearsal can hinder adaptability, and there is a decreasing floor in memory error under the overfitting regimes.
    \item We validate these theoretical findings through numerical simulations and futher extend them with experiments  using DNNs on multiple real-world datasets.
    \end{itemize}

\section{Related Work}
\label{section2_related Work}

    Prior research in continual learning has primarily focused on empirical studies. These methods can be broadly categorized into three main categories: rehearsal-based methods \cite{replay1GEM2017,replay2nature2020,replay5neurps2024}, where a portion of the previous task data is stored and replayed to mitigate forgetting while learning new tasks. Expansion-based methods \cite{structure1PNN2016,structure2tnnls2022,structure3cvpr2022,structure6transip2024} allocate separate network parameters to learn tasks without interfering with previously learned ones, and recent research leverages pre-trained models with lightweight adaptations such as prompting or adapters \cite{pretrain2IJCV2025,pretrain4ICCV2023, pretrain3CVPR2022,pretrain1IJCAI2024}. Regularization-based methods \cite{zhengzehua1EWC2017,zhengzehua3iclr2021,zhengzehua4cvpr2023} constrain the parameters crucial to previous tasks when learning new tasks.

    The theoretical studies in continual learning mainly focus on elucidating its dynamic evolutionary mechanisms through diverse frameworks and establishing links to related domains \cite{1survey3tpami2024,kseparabledata2023,kstatisticaltheory2024}. \cite{ktheoreticalstudyood2022} reformulates continual learning as a combination of within-task prediction and task-id prediction, with links to out-of-distribution detection. \cite{kidealcontinual2023} proposes a general formulation of ideal continual learning, linking it to related areas and providing generalization bounds for replayed samples. \cite{kforgetgenerate2023} focuses on regularization-based continual learning estimators and demonstrates the phenomenon of benign overfitting in continual learning. Connections between continual learning and alternating projections or Kaczmarz methods have also been explored, leading to worst-case forgetting bounds \cite{khowcatastrophic2022}. The trade-off between forgetting and generalization is modeled as a two-player game solved via dynamic programming \cite{ktwoplayergame2021}. \cite{kfixdesign2023} studies continual ridge regression with non-random features, focusing on the impact of regularization.

    Recent studies have investigated rehearsal-based continual learning from multiple perspectives, examining the effects of rehearsal strategies\cite{compare2arxiv2025}, sampling schemes\cite{compare3arxiv2025}, step sizes\cite{compare4arxiv2024}, and network width\cite{compare5icais2023}. \cite{compare1mubiaoarxiv2024} analyzed the role of model dimensionality, highlighting the potential benefits of increased model dimensionality for both multi-task and continual learning. \cite{compare2arxiv2025} compared concurrent and sequential rehearsal strategies, showing that sequential rehearsal outperforms concurrent rehearsal when tasks differ substantially. In contrast to prior work, we focus on the impact of rehearsal scale, a critical yet underexplored factor, and uncover a series of intriguing and counterintuitive findings. Specifically, we show that increasing rehearsal scale can impair model adaptability, a conclusion that contrasts sharply with the conventional view. We validate these theoretical insights through numerical simulations and extensive experiments with deep neural networks on multiple real-world datasets, revealing the double-edged effect of rehearsal. These theoretical findings provide new perspectives into the role of rehearsal scale in continual learning.

\section{Preliminaries}
\label{section3_preliminaries}

    \textbf{Data.} \; 
     We consider a standard continual learning problem where tasks are introduced sequentially, indexed by $t=1,2,...,T$. Suppose that each task $t$ holds a dataset $\mathcal D_{t}= \left \{  (x_{t,i},y_{t,i})\in \mathbb{R}^{p}\times \mathbb{R}\right \}_{i=1}^{n_t}$, where $n_t$ denotes its sample size. Here, $x_{t,i}$ denotes the feature vector and $y_{t,i}$ denotes the corresponding response variable. Each pair $(x_{t,i},y_{t,i})$ follows model $y_t=\boldsymbol x_t^{\top }\boldsymbol w_t^{*}+\epsilon_t$, where $\epsilon_t$ is random noise and $\boldsymbol w_t^{*}$ represents the optimal parameter of the $t$-th task specific model. The equation above can be rewritten into a compact matrix equation for training samples:
    \begin{equation}
        \boldsymbol y_t =\boldsymbol X_t ^{\top} \boldsymbol w_t^{*}+\boldsymbol \epsilon_t ,
    \end{equation}
    where the matrix $\boldsymbol X_t:= [x_{t,1},x_{t,2},...,x_{t,n_t}]\in \mathbb R^{p\times n_t}$, $\boldsymbol y_t:= [y_{t,1},y_{t,2},...,y_{t,n_t}]^{\top }\in \mathbb R^{n_t}$, and $\boldsymbol \epsilon_t:= [\epsilon_{t,1},\epsilon_{t,2},...,\epsilon_{t,n_t}]^{\top }\in \mathbb R^{n_t}$.
    For analytical tractability, the Gaussian features and noise are adopted in this setup.

    \begin{assumption} 
    \label{assumption1}      
         For $t\in T$, element of $\boldsymbol X_t$ follows i.i.d standard Gaussian. The noise $\boldsymbol \epsilon_t$ is independently drawn from $N(0,\sigma_t ^2I_{n_t})$, where $\sigma_t \ge 0$ denotes noise level.
    \end{assumption}	  

    \begin{remark}
        The Gaussian setup is used to exploit the properties of orthogonal projection inspired by \cite{kforgetgenerate2023,normal1NIPS2023,normal2ICLR2025}. 
    \end{remark}

    \begin{assumption} 
    \label{assumption2}      
     The samples $n_t=n$, $\sigma_t=\sigma $ for $t\in T$.
    \end{assumption}	

    \begin{remark}
         Note that analyzing this model provides a critical first step toward understanding deep neural networks, as shown in recent studies \cite{khowcatastrophic2022,linear1JMLR2023}, and our results are further validated through DNNs on real-world datasets with longer task sequences and deeper architectures in Section~\ref{section5_dnn experiments}.
    \end{remark}

    \graphicspath{{muni/}}
    \begin{figure*}[!t]
        \centering
        \begin{minipage}[t]{0.32\textwidth}
            \centering
            \includegraphics[width=\linewidth]{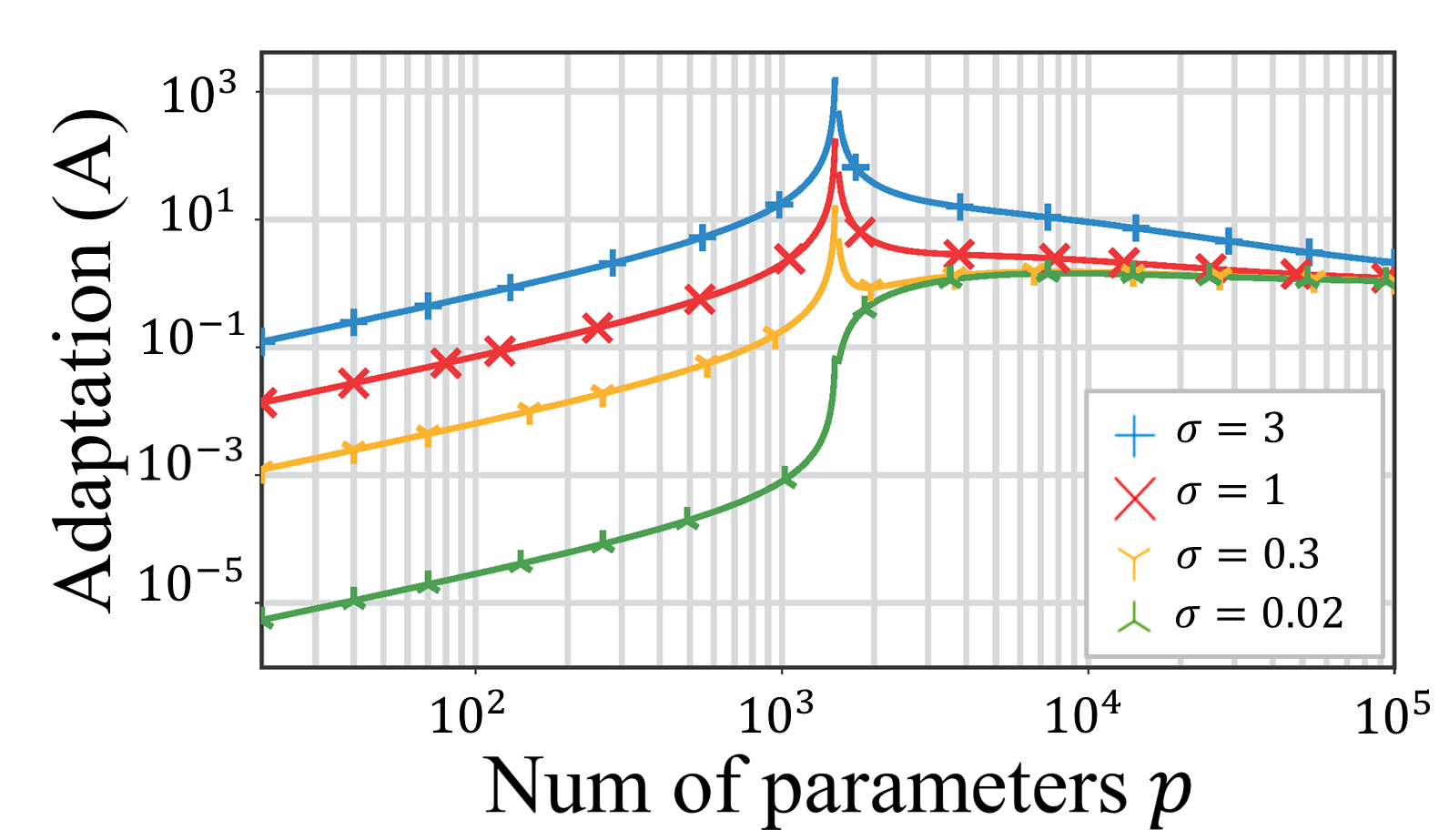}\\
            (a)
        \end{minipage}
        \hfill
        \begin{minipage}[t]{0.32\textwidth}
            \centering
            \includegraphics[width=\linewidth]{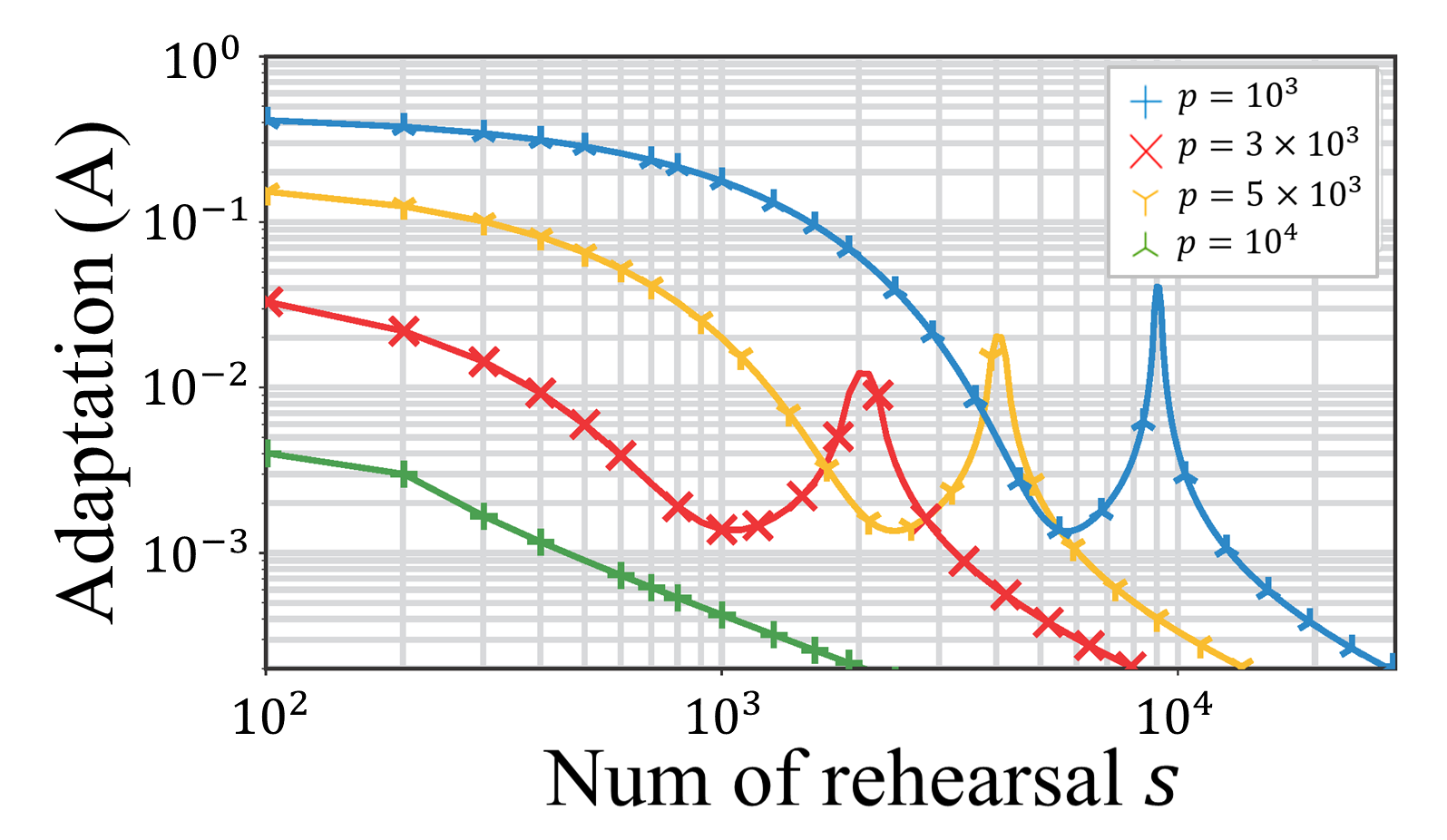}\\
            (b)
        \end{minipage}
        \hfill
        \begin{minipage}[t]{0.32\textwidth}
            \centering
            \includegraphics[width=\linewidth]{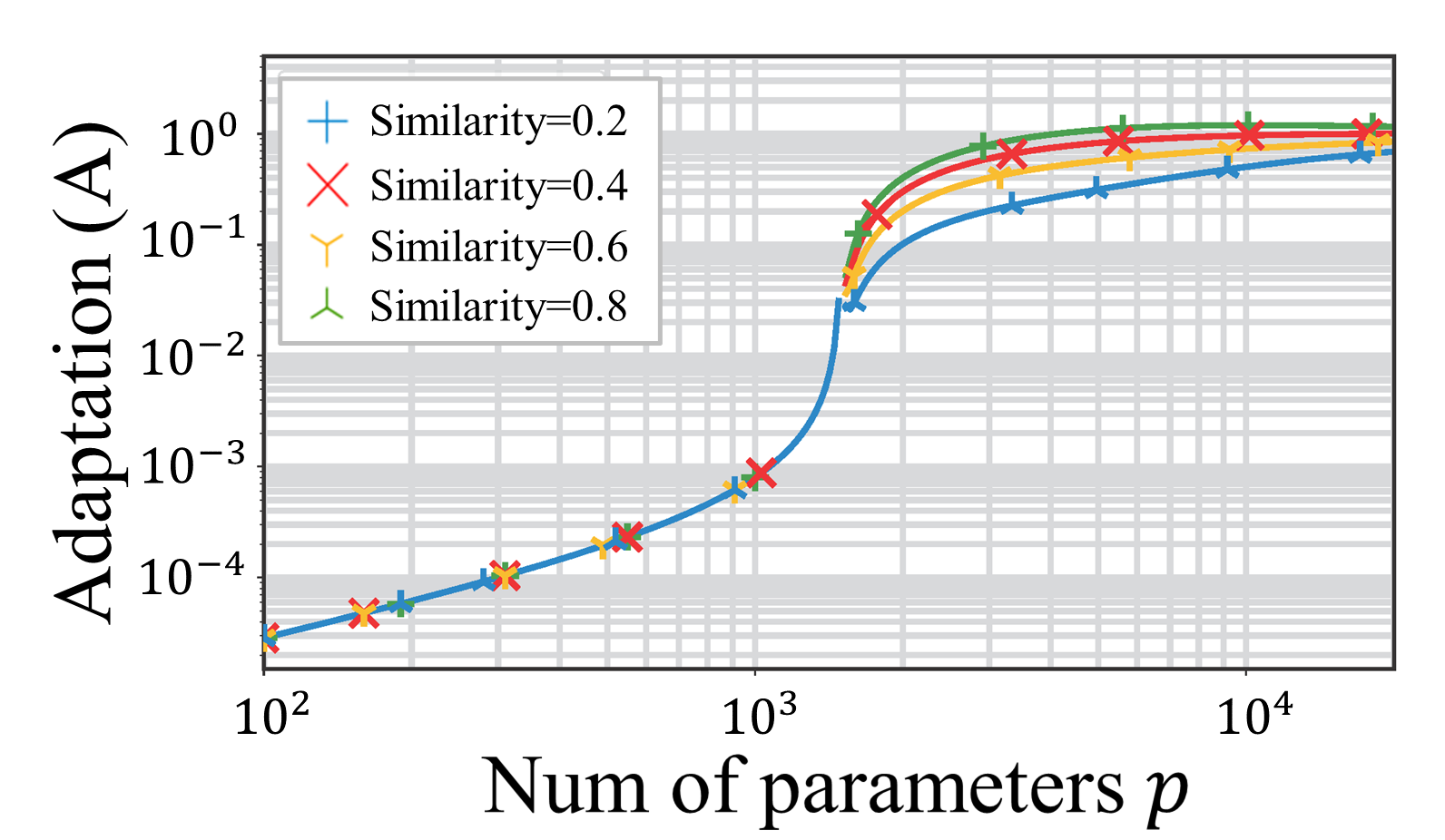}\\
            (c)
        \end{minipage}
        \caption{
        Adaptation error of rehearsal-based methods under different setups,
        where $T=8$, $n=1000$ and $\|\boldsymbol w_t^{*}\|^2=1$ for all $t\in T$.
        Subfigure settings: (a) $s=500$; (b) $\sigma=0.02$; (c) $s=500$, $\sigma=0.02$. Discrete points denote averages across runs.  }
        \label{图片_EAT}
    \end{figure*}

    \textbf{Evaluation metrics.} \; 
     For estimator $\hat{\boldsymbol w}$, we denote its estimation error by $\mathcal L(\hat{\boldsymbol w})=\left \| \hat{\boldsymbol w}-\boldsymbol w^* \right \| ^2$. Based on $\mathcal L(\hat{\boldsymbol w})$, the adaptation error ($\mathcal{A}$) \cite{1survey3tpami2024}, generalization error ($\mathcal{G}$) \cite{kforgetgenerate2023} and memory error ($\mathcal{M}$) \cite{kforgetgenerate2023} can be defined as follows 
     \begin{align}  
     &\mathcal{A}\left(\hat{\boldsymbol{w}_t}\right):= \left \| \hat{\boldsymbol w_{t}}-\boldsymbol w_{t}^* \right \| ^2 ,   \\
    & \mathcal{G}\left(\hat{\boldsymbol{w}_t}\right):= \frac{1}{t}  \sum_{i=1}^{t}  \left \| \hat{\boldsymbol w_{t}}-\boldsymbol w_{i}^* \right \| ^2 ,      \\ 
    &\mathcal{M}\left(\hat{\boldsymbol{w}_t}\right):= \frac{1}{t-1}  \sum_{i=1}^{t-1} [  \left \| \hat{\boldsymbol w_{t}}-\boldsymbol w_{i}^* \right \| ^2 - \left \| \hat{\boldsymbol w_{i}}-\boldsymbol w_{i}^* \right \| ^2 ] , 
    \end{align}  
     for each $t\in T$, where $\hat{\boldsymbol w_{t}}$ denotes the parameters after task t has been learned. Due to the parameter differences reflect functional differences, the distance between optimal parameters is defined to measure inter-task similarity. A algorithm obtains increasing performance on previous tasks if, for each $t\in T$, the forgetting error satisfies $\mathcal{M}\left(\hat{\boldsymbol{w}_t}\right)<0$.

    \textbf{Rehearsal-based Continual Learning Estimator.} The rehearsal-based continual estimator assumes tasks arrive sequentially and preserves knowledge of previous tasks by storing a subset of their samples \cite{1survey1nn2019,1survey2tpami2021,estimatereplay1neurps2018,estimatereplay2arxiv2022,estimatereplay3arxiv2023}. For each task $t=2,...,T$, assume that a total of $s$ samples are stored. Specifically, we assume that the feature vector matrix of the $i$-th previous task ( $i=1,2,...,t-1$ ) stored in the memory buffer is $\boldsymbol Z_i\in \mathbb R^{p\times \frac{s}{t-1}} $, with the corresponding response variable denoted as $\boldsymbol g_i\in \mathbb R^{\frac{s}{t-1} } $. The training process converges to the optimal solution by minimizing the training loss, formulated as the following optimization problem: $\widehat{\boldsymbol{w}}_{t}^{(\text {Reh})} := \underset{\boldsymbol{w}}{\arg \min } \left\|\boldsymbol{X}_{t}^{\top} \boldsymbol{w}-\boldsymbol{y}_{t}\right\|^{2}  + \sum_{i=1}^{t-1} \left\|\boldsymbol{Z}_{i}^{\top} \boldsymbol{w} -\boldsymbol{g}_{i}\right\|^{2} $. When $p>n+s$ (overparameterized), multiple solutions exist that achieve zero training loss. In this case, we select the solution with the minimum $\ell_2$-norm , i.e., the optimization problem: 
    $\arg\min_{\boldsymbol{w}} \bigl\{ \|\boldsymbol w - \boldsymbol w_{t-1}\|^{2}, \text{ s.t. } (\boldsymbol X_{t})^\top \boldsymbol w = \boldsymbol y_{t}, (\boldsymbol Z_{i})^\top \boldsymbol w = \boldsymbol g_{i}, i = 1,\dots,t-1 \bigr\}$. Among all overfitting solutions, we focus on the minimum $\ell_2$-norm solution because it corresponds to the convergence point of stochastic gradient descent in continual learning. In Section~\ref{section4_main results}, we provide theoretical results for rehearsal-based learner.

    \textbf{The impact of rehearsal scale on adaptability, memorability, and generalization.} Unlike prior studies that focused on other foundational factors, we theoretically analyze how rehearsal scale affects performance. The adaptation error measures fit to the current task, the memory error captures performance loss on previous tasks, and the generalization error evaluates transfer to unseen tasks. Section~\ref{section4_main results} explores the relationship between rehearsal scale and these three aspects.

\section{Main Results}
\label{section4_main results}

    In this section, we present the main results. For rehearsal-based continual learning, we establish three theorems that characterize adaptation, memory and generalization error under overparameterized and underparameterized regimes.

    \begin{theorem}[Adaptation error]
    \label{theorem1}
    Suppose that Assumption~\ref{assumption1}-\ref{assumption2} hold. Then the adaptation error of the rehearsal-based continual learning is formally given by:  \\
    under the overparameterized regime, there have
    \begin{align}
    \mathbb{E}\!\left[\mathcal{A}(\widehat{\boldsymbol{w}}_{T})\right]
    &=
    \lambda^{T} \left\|\boldsymbol{w}_{T}^{*}\right\|^{2}
    + a_{\text{noise}}  \notag  \\ 
    &+ \underbrace{ \sum_{k=1}^{T} \lambda^{T-k} \frac{n+s}{p}
    \left\|\boldsymbol{w}_{k}^{*}-\boldsymbol{w}_{T}^{*}\right\|^{2} }_{\text{Term A1}},
    \end{align}   
     under the underparameterized regime, there have
    \begin{align}
    \mathbb{E}\!\left[\mathcal{A}(\widehat{\boldsymbol{w}}_{T})\right]
    =
    \frac{p\sigma^{2}}{n+s-p-1},
    \end{align}
    where $\lambda := \frac{p-n-s}{p} $ and $a_{noise}:=\frac{(1-\lambda^{T})p\sigma^{2}}{(p-n-s-1)} $, with larger $\lambda$ indicating greater overparameterization.
    \end{theorem}

    The proof is provided in Appendix B. It describes the ability to fit current task, forming the basis for analyzing how it learns new knowledge and retains previous knowledge.

    \textbf{Increasing the rehearsal size enhances the model’s adaptability under underparameterization, whereas it can be detrimental under overparameterization.} Specifically, when $n+s>p+1$ in Equation (6), $\mathbb{E}[\mathcal A (\widehat{\boldsymbol{w}}_{T} )]$ decreases as $s$ increases, indicating that more playback contributes to better adaptation. When slightly overparameterized in Equation (5), we have $p\approx n+s$ and thus $\lambda \approx 0$. At this point, Term A1 and the denominator in Term $a_{noise}$ approach zero when tasks are similar, and thus $a_{noise}$ dominates and causes $\mathbb{E}[\mathcal A (\widehat{\boldsymbol{w}}_{T} )]$ to be increasing w.r.t. $s$.  When heavily overparameterized in Equation (5), Term A1 is close to zero, and thus $\mathbb{E}[\mathcal A (\widehat{\boldsymbol{w}}_{T} )]$ decreases as $s$ increases when $\sigma$ is low. Intuitively, when tasks are similar, the model can leverage rehearsal more effectively, leading to improvements in performance on current task.

    \graphicspath{{muni/}}
    \begin{figure*}[t]
        \centering
        \begin{minipage}[t]{0.32\textwidth}
            \centering
            \includegraphics[width=\linewidth]{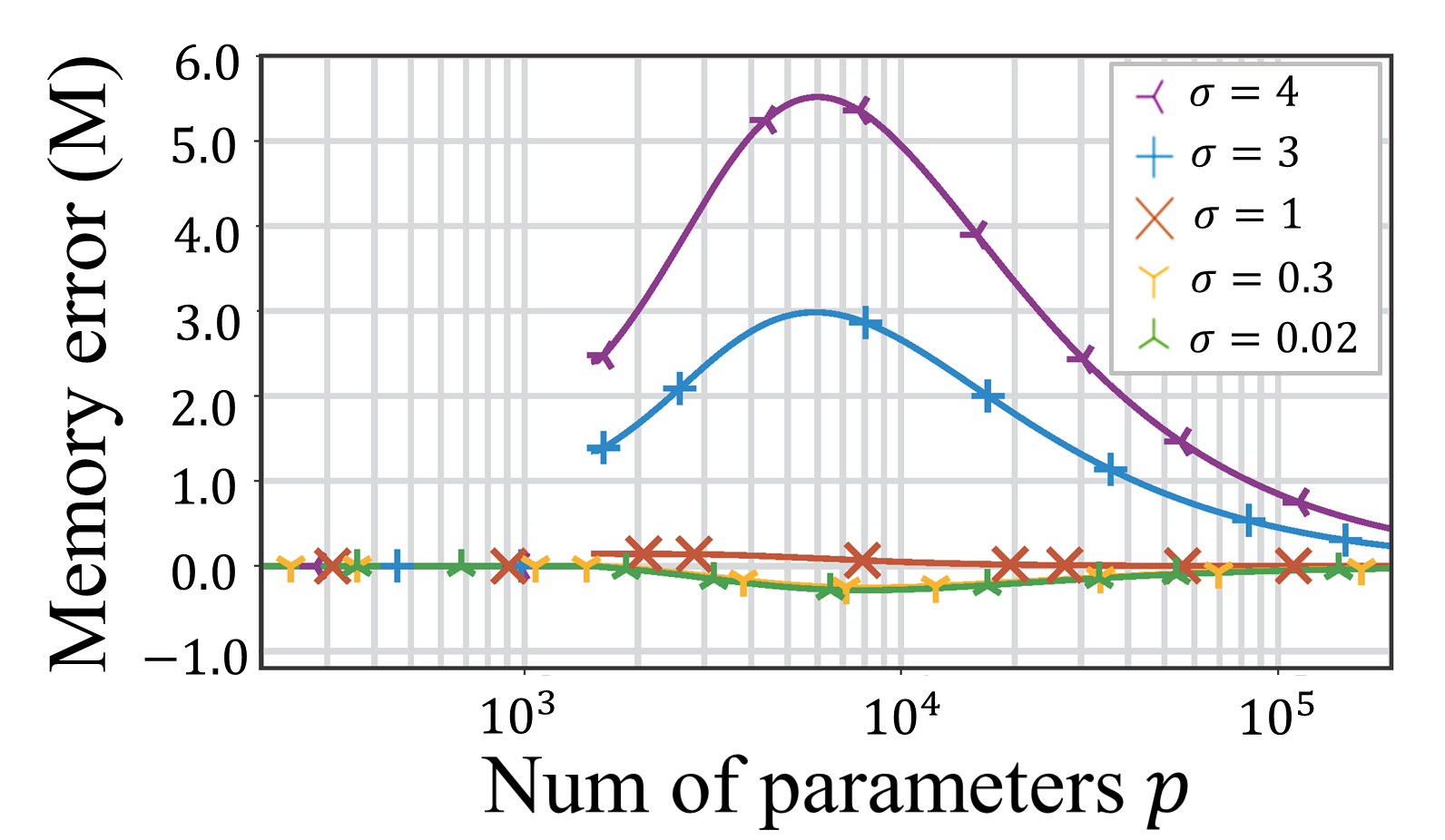}\\
            (a)
        \end{minipage}
        \hfill
        \begin{minipage}[t]{0.32\textwidth}
            \centering
            \includegraphics[width=\linewidth]{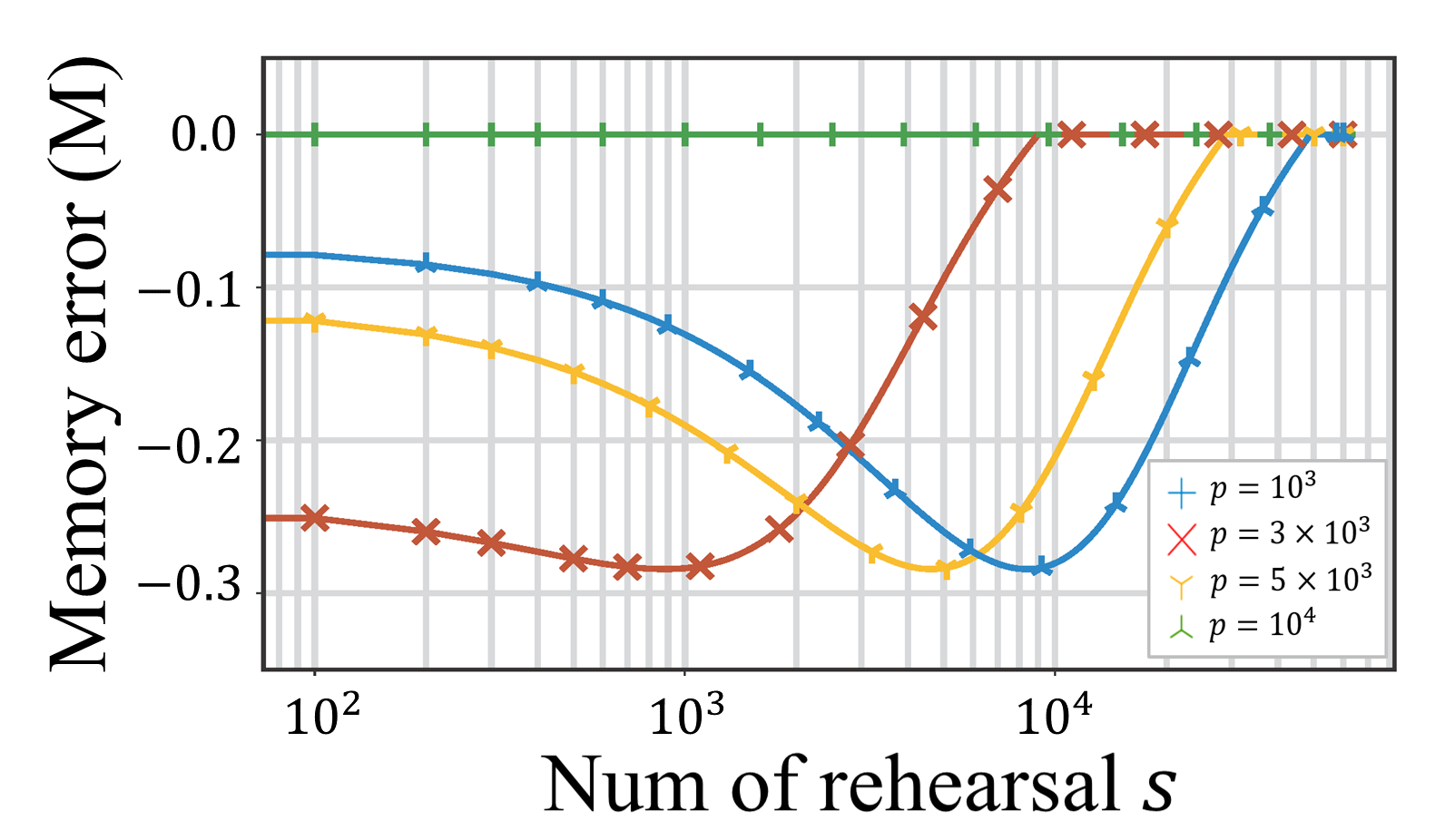}\\
            (b)
        \end{minipage}
        \hfill
        \begin{minipage}[t]{0.32\textwidth}
            \centering
            \includegraphics[width=\linewidth]{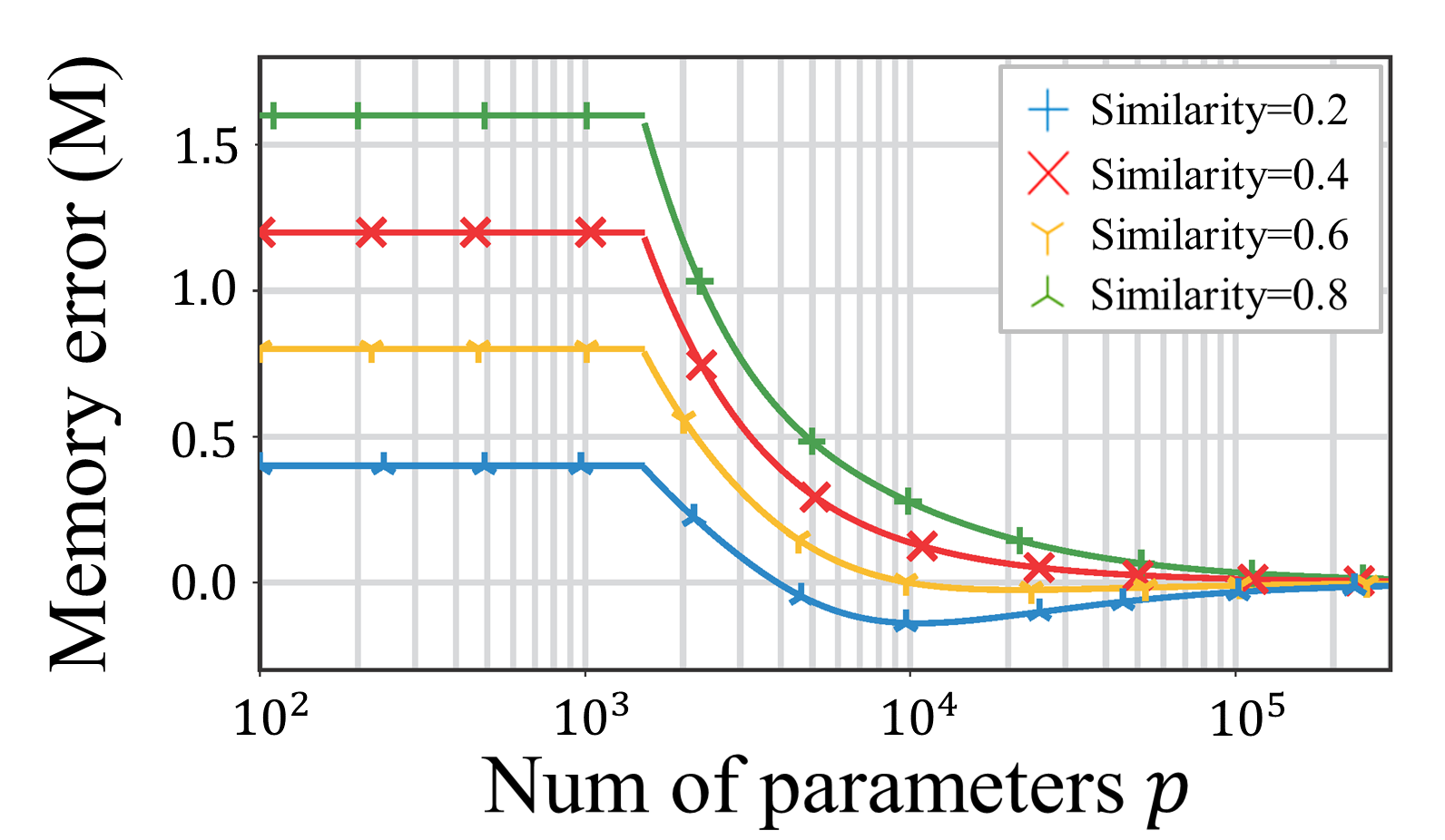}\\
            (c)
        \end{minipage}
        \caption{
        Memory error of rehearsal-based methods under different setups,
        where $T=8$, $n=1000$ and $\|\boldsymbol w_t^{*}\|^2=1$ for all $t\in T$.
        Subfigure settings: (a) $s=500$; (b) $\sigma=0.02$; (c) $s=500$, $\sigma=0.02$. Discrete points denote averages over simulation runs. }
        \label{图片_EMT}
    \end{figure*}

    The impact of rehearsal size was also verified through numerical simulations in Figure~\ref{图片_EAT}(b), where average adaptation error is plotted against rehearsal size for different model parameters. The red curve marked with "$\times$", decreases in the underparameterized regime ($s>p-n$) but first decreases and then increases in the overparameterized regime ($s<p-n$), which validates our earlier insights.

    \begin{remark}(Turning point.)
        Based on the above analysis, we characterize the inflection point under overparameterization by computing the partial derivative with respect to $s$. Specifically, when $t=8$, $n=1000$, $p=3000$ and $\sigma=0.02$, the inflection point of $s$ is calculated to be near $1000$, consistent with the red curve shown in Figure~\ref{图片_EAT}(b).
    \end{remark}

    Beyond analyzing the impact of rehearsal size, we also examined the effects of inter-task similarity and model parameters. We found that under overparameterization, models require higher inter-task similarity to better adapt to the current task, whereas this effect is not observed in underparameterized settings (Figure~\ref{图片_EAT}(c)). Moreover, the overparameterization helps mitigate the impact of task variability and noise effects on model's adaptability, as illustrated in Figure~\ref{图片_EAT}(a) and Figure~\ref{图片_EAT}(c). Due to space constraints, a more detailed discussion of these factors is provided in Appendix G.

    \begin{theorem}[Memory error]
    \label{theorem2}
    Under Assumption~\ref{assumption1} and Assumption~\ref{assumption2}, the memory error of the rehearsal-based continual learning model is formally given by 
    
     under the overparameterized regime, there have
    \begin{equation}
    \begin{aligned}
    \mathbb{E}[\mathcal M(\widehat{\boldsymbol{w}}_{T})]
    =&\;
    \underbrace{\frac{1}{T-1} \sum_{k=1}^{T-1}\sum_{j>k}^{T}
    \frac{n+s}{p}u_{kj}
    \left\|\boldsymbol{w}_{j}^{*}-\boldsymbol{w}_{k}^{*}\right\|^{2}}_{\text{Term M1}} \\
    &+
    \underbrace{\frac{1}{T-1} \sum_{i=1}^{T-1}
    \left( \lambda^{T}-\lambda^{i}\right)
    \left\|\boldsymbol{w}_{i}^{*}\right\|^{2}}_{\text{Term M2}}
    + m_{\text{noise}},
    \end{aligned}
    \end{equation}
    
    under the underparameterized regime, there have
    \begin{equation}
    \mathbb{E}[\mathcal M(\widehat{\boldsymbol{w}}_{T})]
    = \frac{1}{T-1} \sum_{k=1}^{T-1}
    \left\|\boldsymbol{w}_{T}^{*}-\boldsymbol{w}_{k}^{*}\right\|^{2},
    \end{equation}
 
    where $u_{kj}:=\lambda^{T-k}-\lambda^{j-k}+\lambda^{T-j}$ and  $m_{noise}:= \frac{1}{T-1} \sum_{i=1}^{T-1}  \frac{p \sigma^{2}}{p-n-s-1} \left( \lambda^{i} - \lambda^{T} \right)$. Specifically, when $T=2$, the above Equation (7) can be reformulated as follows
    \begin{equation}
    \begin{aligned}
         &\mathbb{E}[\mathcal M(T=2)] = - \frac{(n+s)(p-n-s)}{p^2} \left\|\boldsymbol{w}_{1}^{*}\right\|^{2}  \\
        &+ \frac{n+s}{p}\left\|\boldsymbol{w}_{2}^{*}-\boldsymbol{w}_{1}^{*}\right\|^{2} 
         +\frac{(n+s)(p-n-s)\sigma ^2}{(p-n-s-1)p}.    
    \end{aligned}
    \end{equation}
    Similarly, the Equation (8) can be reformulated as 
    \begin{equation}
    \mathbb{E}[\mathcal{M}(T=2)] = \left\|\boldsymbol{w}_{2}^{*}-\boldsymbol{w}_{1}^{*}\right\|^{2}.
    \end{equation}
    \end{theorem}

    The detailed proof is provided in Appendix C. Based on Theorems~\ref{theorem2}, we further explore their analytical insights and examine the influence of factors such as rehearsal size, as well as the performance differences observed under both overparameterized and underparameterized regimes.

    \textbf{Increasing the rehearsal does not always lead to better memorability in continual learning models under the overparameterized regime.} Specifically, we consider the case where $T = 2$. For the overparameterized regime result in Equation (9), when tasks are similar and $\sigma$ is low, the second term dominates and causes $\mathbb{E}[\mathcal M(\widehat{\boldsymbol{w}}_{T} )]$ first decreases and then increases as $s$ increases, indicating the existence of decreasing floor. For the underparameterized regime result in Equation (10), rehearsal no longer contributes to memorability. In this situation, the $\mathbb{E}[\mathcal M(\widehat{\boldsymbol{w}}_{T} )]$ depends solely on the inherent similarity between tasks ( i.e. $\left\|\boldsymbol{w}_{T}^{*}-\boldsymbol{w}_{k}^{*}\right\|$ ), meaning that the memory error is fully determined by similarity between the final task and preceding ones.

    The yellow curve marked "$\text { Y }$" in Figure~\ref{图片_EMT}(b) illustrates how the average memory error varies with rehearsal size when $p=3\times 10^4$. In the overparameterized regime ($s<p-n$), the memory error first decreases and then increases, indicating the existence of a decreasing performance floor. In contrast, in the underparameterized regime ($s>p-n$), the error remains unaffected, and zero forgetting is achieved when the task-optimal parameters remain consistent.

    \graphicspath{{muni/}}
    \begin{figure*}[t]
        \centering
        \begin{subfigure}[t]{0.32\linewidth}
            \centering
            \includegraphics[width=\linewidth]{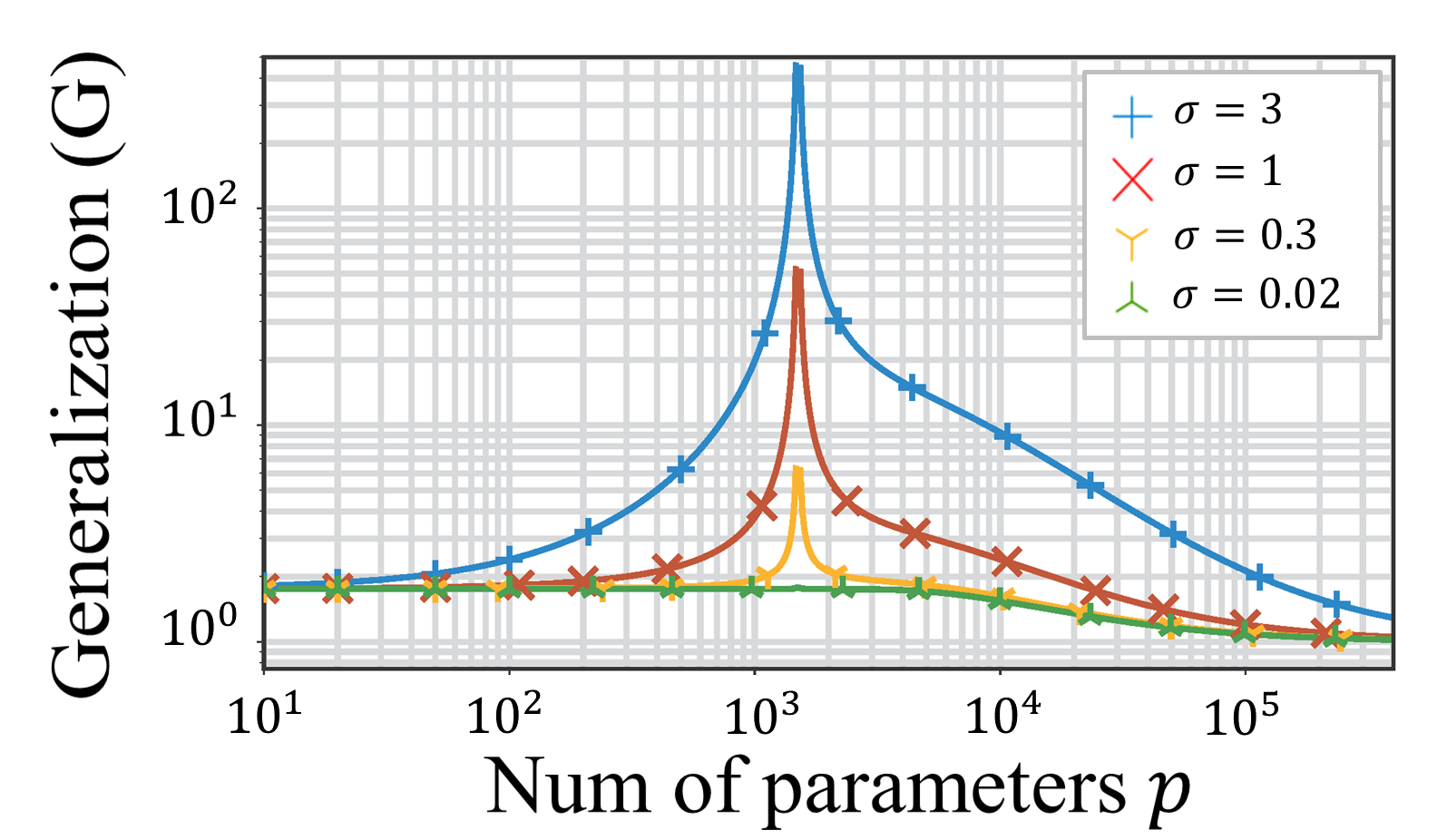}
            \caption{}
        \end{subfigure}
        \hfill
        \begin{subfigure}[t]{0.32\linewidth}
            \centering
            \includegraphics[width=\linewidth]{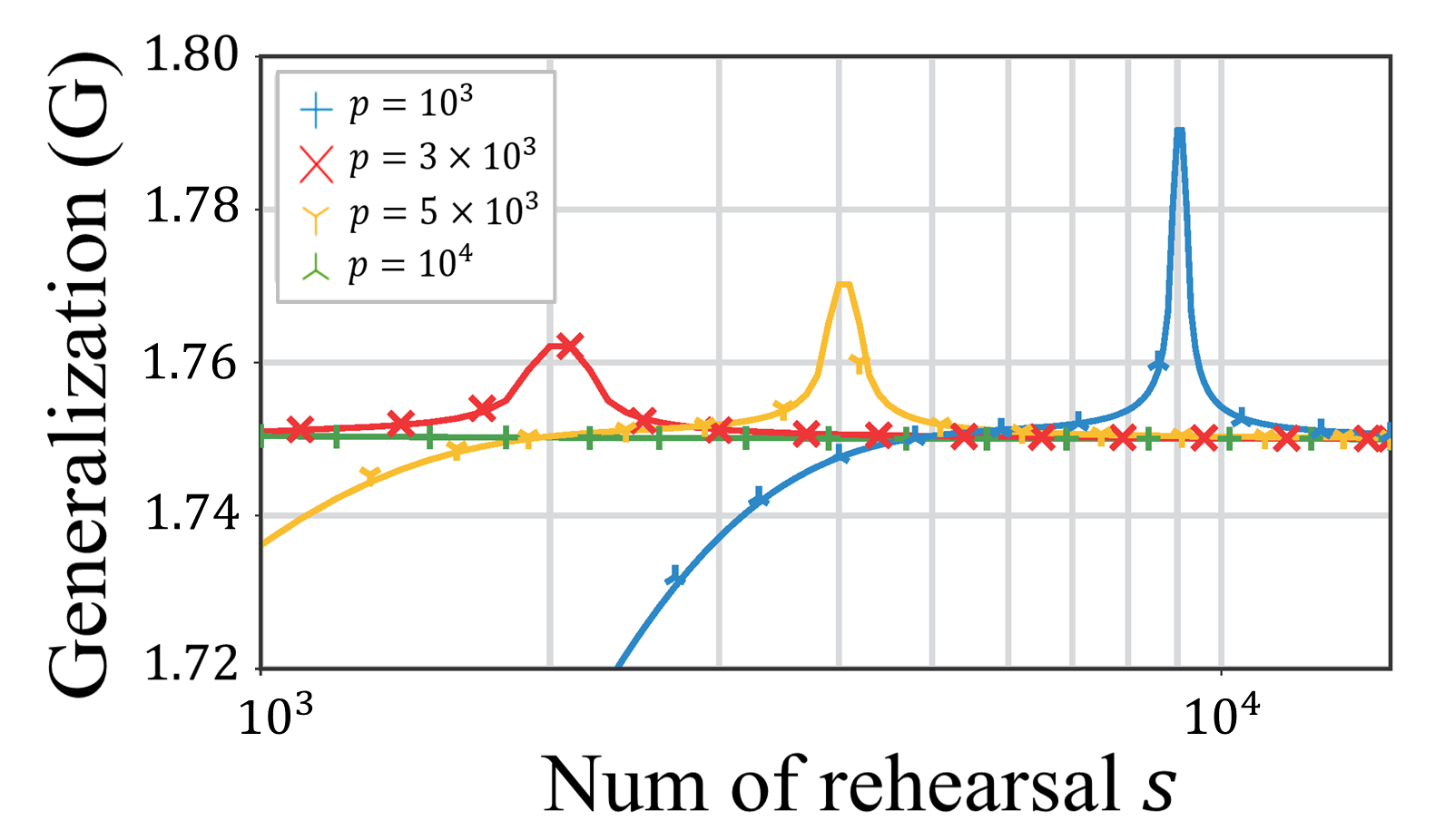}
            \caption{}
        \end{subfigure}
        \hfill
        \begin{subfigure}[t]{0.31\linewidth}
            \centering
            \includegraphics[width=\linewidth]{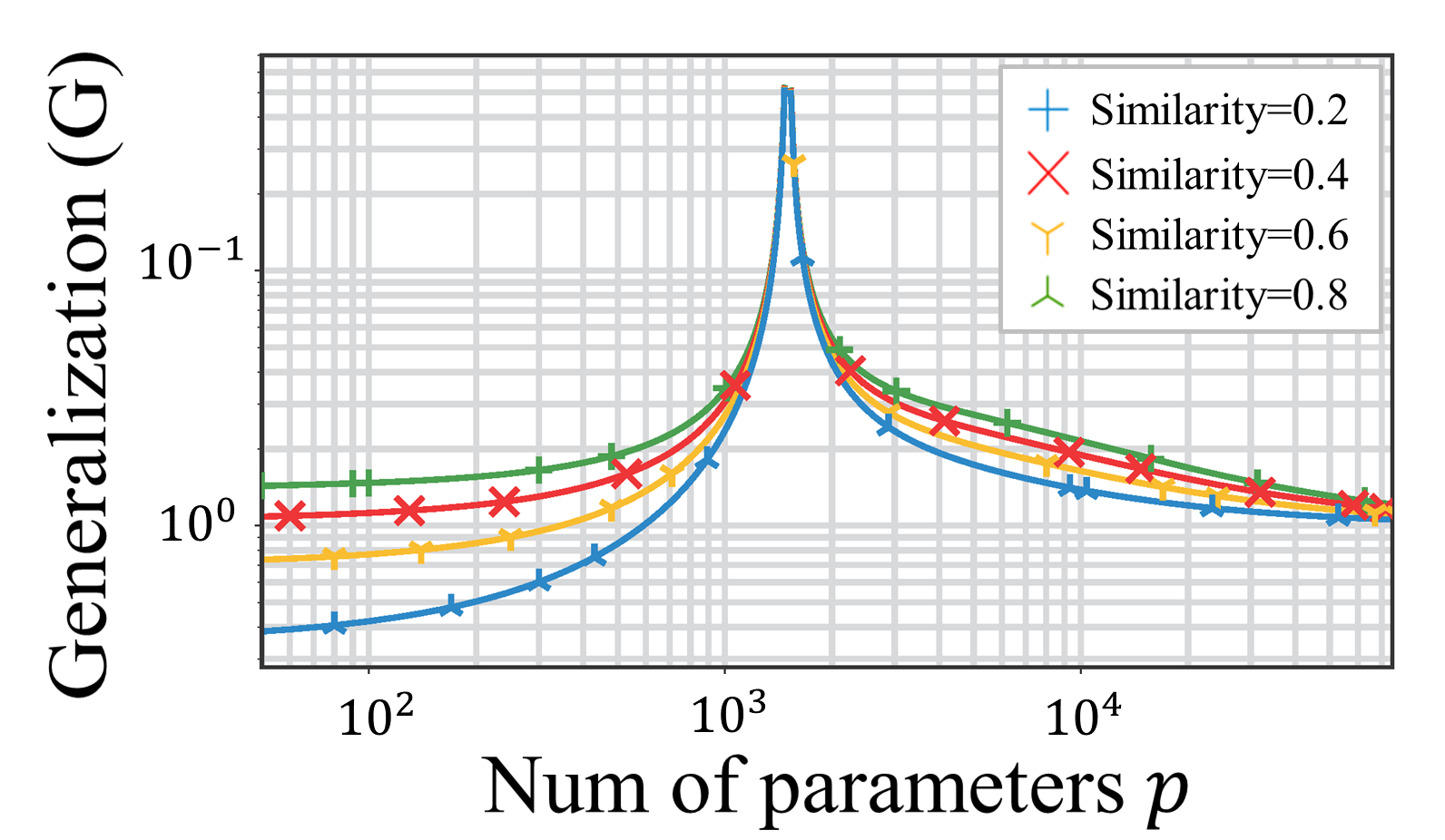}
            \caption{}
        \end{subfigure}
        \caption{
        Generalization error w.r.t. the number of model parameters or rehearsal samples, with $T=8$, $n=1000$ and $\|\boldsymbol w_t^{*}\|^2=1$.
        Subfigure settings: (a) $s=500$; (b) $\sigma=0.02$; (c) $s=500$, $\sigma=0.02$. Discrete points denote averages across runs. }
        \label{图片_EGT}
    \end{figure*}

    \begin{table*}[t]
    \centering
    \normalsize     
    \caption{The adaptation errors on the Tiny-ImageNet dataset across 20 training tasks with varying buffer sizes.}
    \label{表格4_更多任务数量tinyimagenet}
    \begin{tabular}{lccccccc}
    \toprule
    \textbf{Num of Tasks} & \textbf{T3} & \textbf{T6} & \textbf{T9} & \textbf{T12} & \textbf{T16} & \textbf{T18} & \textbf{T20} \\
    \midrule
    0\%  & $1.12 \pm 0.03$ & $0.96 \pm 0.03$ & $1.17 \pm 0.02$ & $0.89 \pm 0.02$ & $0.99 \pm 0.02$ & $1.04 \pm 0.01$ & $1.01 \pm 0.01$ \\
    5\%  & $1.18 \pm 0.02$ & $1.02 \pm 0.03$ & $1.20 \pm 0.03$ & $0.91 \pm 0.02$ & $1.03 \pm 0.00$ & $1.08 \pm 0.01$ & $1.05 \pm 0.02$ \\
    10\% & $1.26 \pm 0.04$ & $1.06 \pm 0.02$ & $1.27 \pm 0.04$ & $0.94 \pm 0.02$ & $1.10 \pm 0.02$ & $1.13 \pm 0.02$ & $1.12 \pm 0.01$ \\
    50\% & $1.45 \pm 0.03$ & $1.22 \pm 0.05$ & $1.42 \pm 0.08$ & $1.09 \pm 0.07$ & $1.21 \pm 0.04$ & $1.31 \pm 0.11$ & $1.27 \pm 0.02$ \\
    \bottomrule
    \end{tabular}
    \end{table*}

    \begin{theorem}[Generalization error]
    \label{theorem3}
    Under Assumption~\ref{assumption1}-\ref{assumption2}, the expected generalization error of the rehearsal-based continual learning model is formally given as follows
    
    under the overparameterized regime, there have
    \begin{equation}
    \begin{aligned}
    \mathbb{E}[\mathcal G(\widehat{\boldsymbol{w}}_{T})]
    =&\;
    \underbrace{\frac{1}{T} \sum_{k=1}^{T}\sum_{j=1}^{T}
    \frac{n+s}{p}\lambda^{T-k}
    \left\|\boldsymbol{w}_{k}^{*}-\boldsymbol{w}_{j}^{*}\right\|^{2}}_{\text{Term G1}} \\
    &+
    \underbrace{\frac{1}{T}\sum_{k=1}^{T}
    \lambda^{T}\left\|\boldsymbol{w}_{k}^{*}\right\|^{2}}_{\text{Term G2}}
    + g_{\text{noise}},
    \end{aligned}
    \end{equation}   
    
    under the underparameterized regime, there have
    \begin{equation}
    \mathbb{E}[\mathcal G(\widehat{\boldsymbol{w}}_{T})]
    =
    \frac{1}{T} \sum_{k=1}^{T}
    \left\|\boldsymbol{w}_{T}^{*}-\boldsymbol{w}_{k}^{*}\right\|^{2}
    + \frac{p\sigma^{2}}{n+s-p-1},
    \end{equation}
    
    where $g_{noise}(\widehat{\boldsymbol{w}}_{T}):= \frac{p \sigma^{2}}{p-n-s-1}\left(1-\lambda^{T}\right)$. 
    Specifically, when $T=2$, we reformulate Equation (11) to provide a clearer interpretation of the error form , resulting in
    \begin{equation}
    \begin{aligned}
        &\mathbb{E}[\mathcal G(T=2)]=
        \frac{1}{2} (1-\lambda^2)\left\|\boldsymbol{w}_{2}^{*}-\boldsymbol{w}_{1}^{*}\right\|^{2}  \\
        &+ \frac{1}{2}\lambda^2
        (\left\|\boldsymbol{w}_{1}^{*}\right\|^{2}+\left\|\boldsymbol{w}_{2}^{*}\right\|^{2})
        +\frac{p\sigma ^2(1-\lambda^2)}{p-n-s-1},  
    \end{aligned}
    \end{equation}

    Similarly, the Equation (12) can be reformulated as
    \begin{equation}
    \mathbb{E}[\mathcal G( T=2 )]= \frac{1}{2} \left\|\boldsymbol{w}_{2}^{*}-\boldsymbol{w}_{1}^{*}\right\|^{2}+\frac{p\sigma ^2}{n+s-p-1}.  \tag{14}
    \end{equation}
    \end{theorem}

    \textbf{Increasing rehearsal can degrade generalization under overparameterization, especially when tasks are dissimilar.} Consider the case where $T=2$. When slightly overparameterized in Equation (13), the second term approaches zero, the denominator $p-n-s-1$ in the third term approaches zero, and thus this term dominates. In this situation, increasing rehearsal size raises the peak generalization error. In contrast, when heavily overparameterized in Equation (13), the second term dominates and decreases as $s$ increases when tasks are similar and $\sigma$ is low. Moreover, the $\mathbb{E}[\mathcal G(\widehat{\boldsymbol{w}}_{T} )]$ decreases as $s$ increases in Equation (14), indicating that larger rehearsal consistently enhances generalization under underparameterization.

    In Figure~\ref{图片_EGT}(b), average generalization error varies with rehearsal size, with optimal parameters being orthogonal. The yellow curve with markers “$\text { Y }$” decreases when underparameterized ($s > p-n$), but increases with rehearsal size when overparameterized ($s < p-n$), confirming these insights. Additionally, we examined other factors under different parameterization regimes. As shown in Figure~\ref{图片_EGT}(c), increasing similarity enhances generalization in both underparameterized and overparameterized settings. However, enlarging the parameter size reduces the influence of rehearsal and inter-task similarity on generalization (Figures~\ref{图片_EGT}(a), \ref{图片_EGT}(c)).

    Noted that the logarithmic axes are used to capture variations in error under different parameterization regimes in Figures~\ref{图片_EAT}–\ref{图片_EGT}. The horizontal axis illustrates the transition from underparameterization to overparameterization, allowing an intuitive comparison. It is worth noting that logarithmic scaling provides a wider coordinate range while preserving the original trends. Similar configurations have been used in \cite{khowcatastrophic2022,kfixdesign2023,kstatisticaltheory2024} to reveal subtle variations in the average error.

    \graphicspath{{picture/}}
    \begin{figure}[H]
        \centering
        \includegraphics[width=\linewidth]{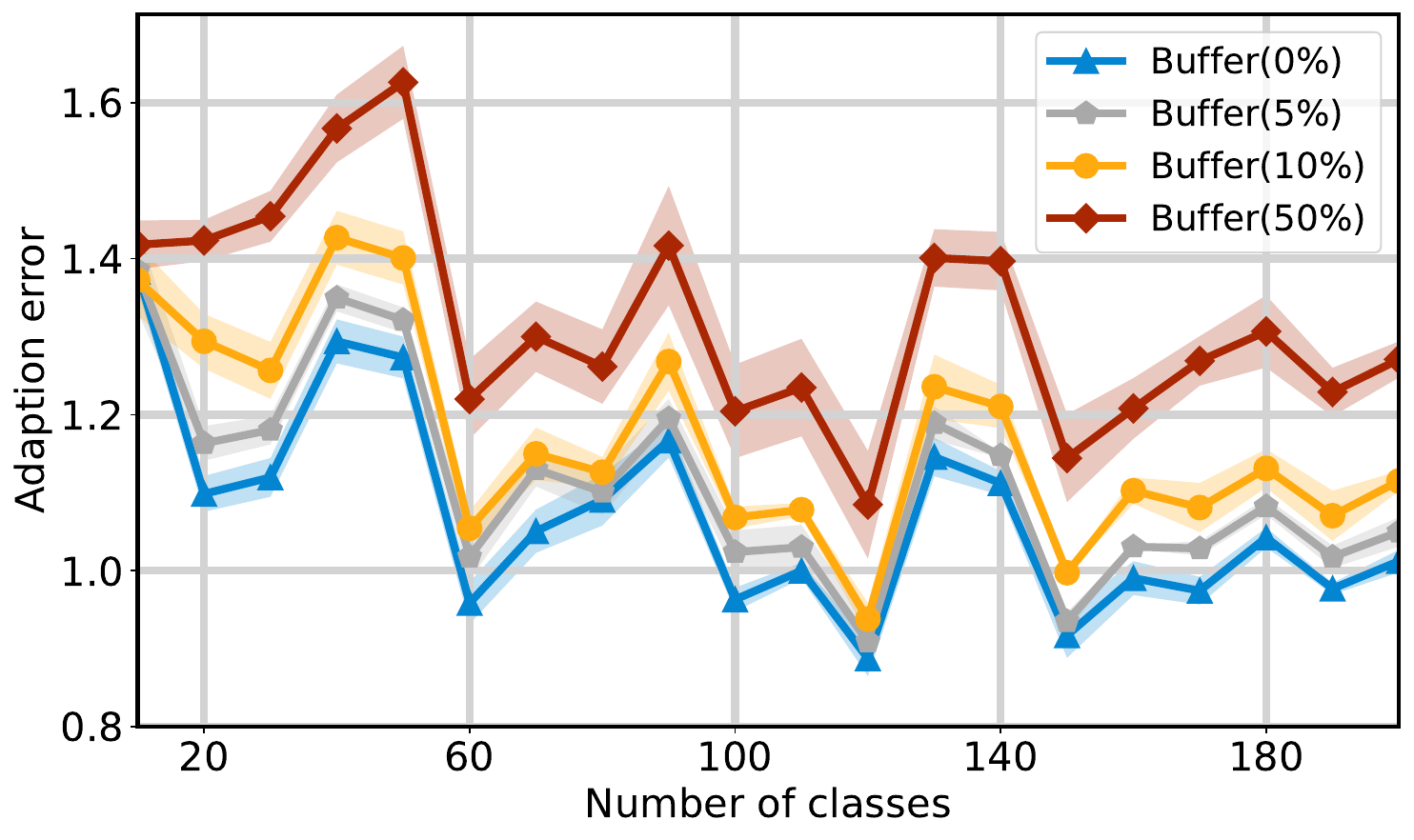}
        \caption{Adaptation error on Tiny-ImageNet with increasing training classes, with the legend showing varying buffer sizes.}
        \label{图片_DNN_tinyimagenet1eat}
    \end{figure}

    In conclusion, we theoretically characterize the effects of rehearsal size on the three types of errors. The following proposition reveals the connection among adaptability, memorability and generalizability for $T=2$, and presents conditions for effective generalization performance.

    \graphicspath{{picture/}}
    \begin{figure*}[t]
        \centering
        \begin{minipage}[t]{0.32\textwidth}
            \centering
            \includegraphics[width=\linewidth]{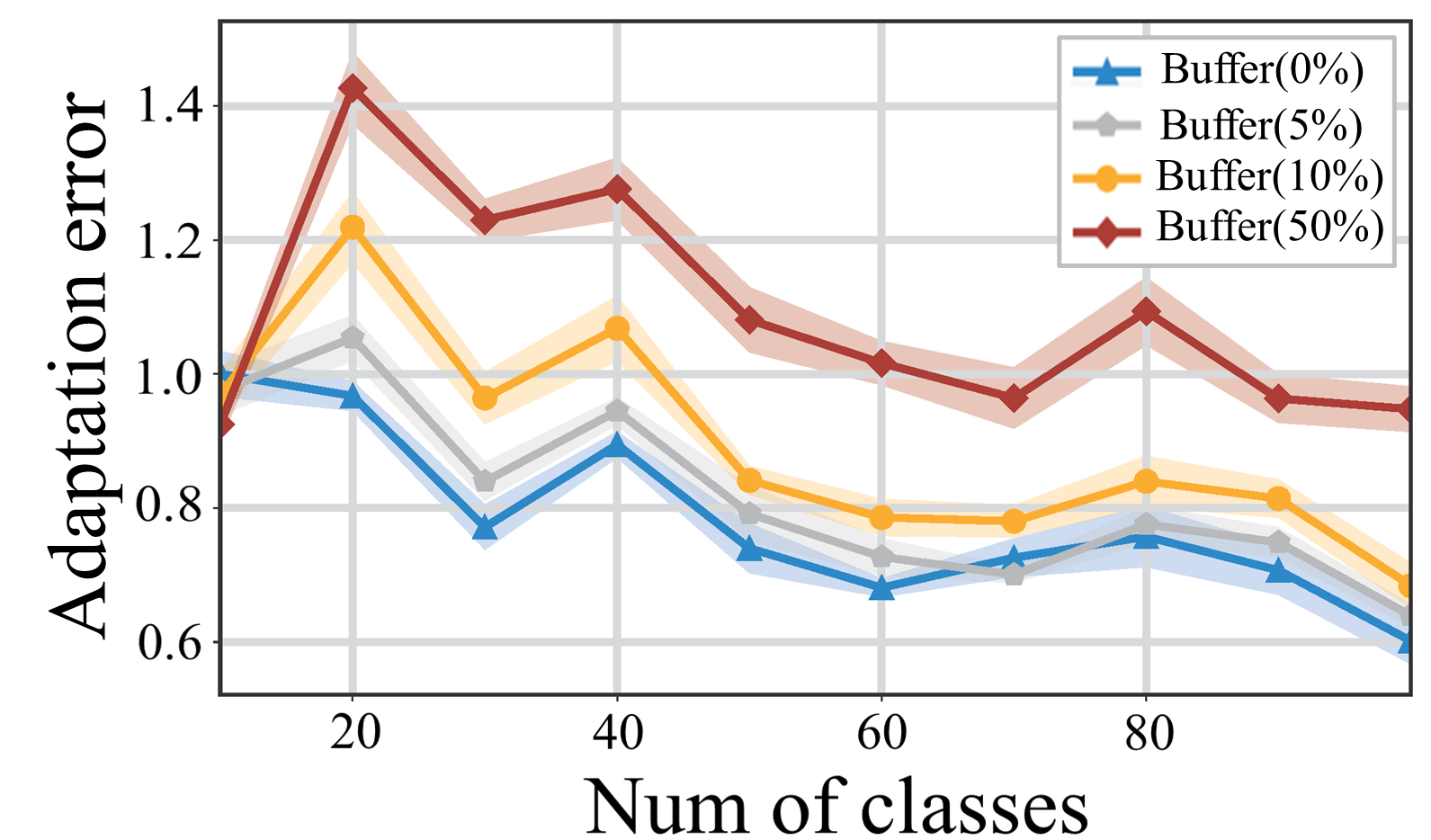}\\
            {\small (a) Adaptation error on CIFAR-100}
            \vspace{4pt}
            
            \includegraphics[width=\linewidth]{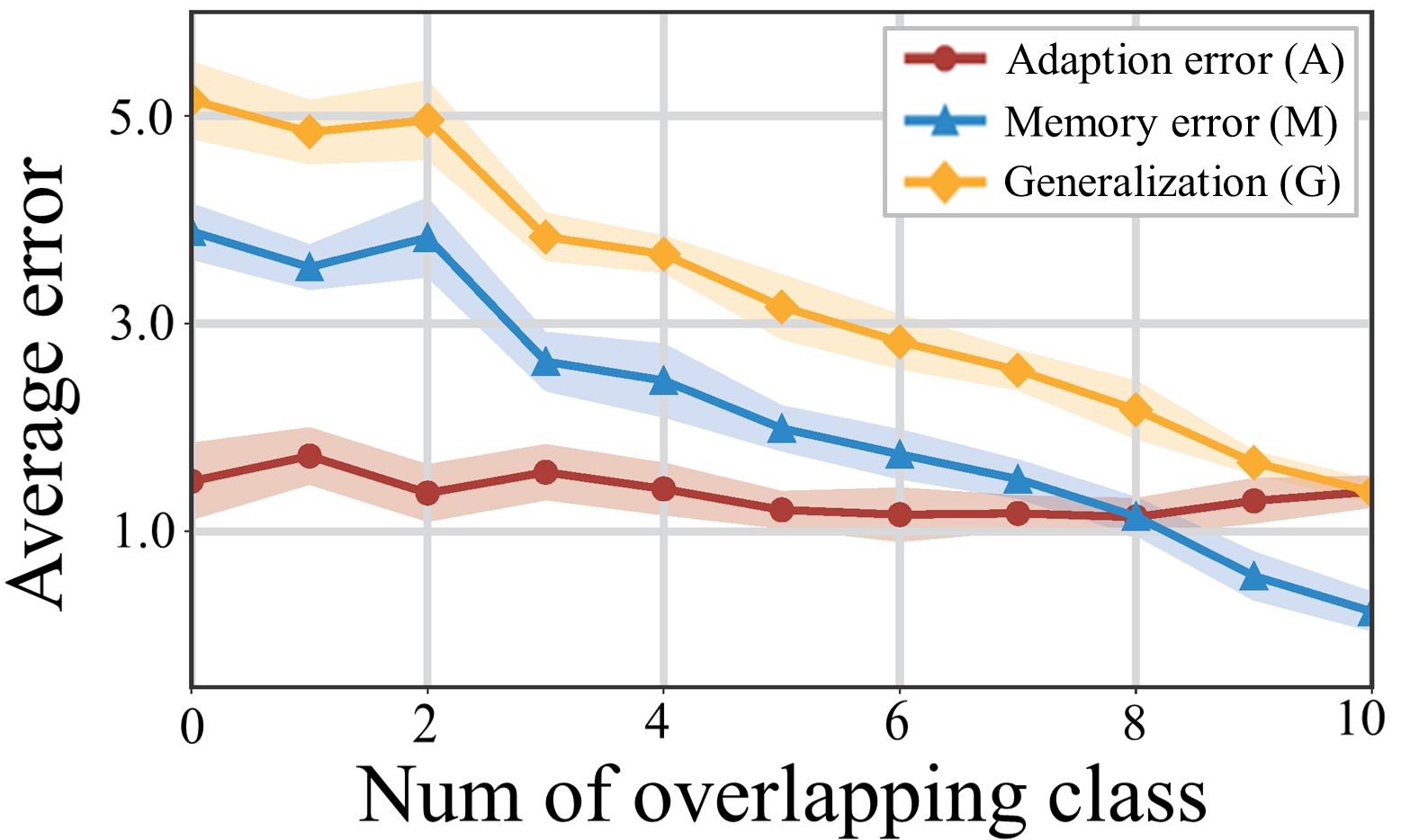}\\
            {\small (d) Average Error on CIFAR-100}
        \end{minipage}
        \hfill
        \begin{minipage}[t]{0.32\textwidth}
            \centering
            \includegraphics[width=\linewidth]{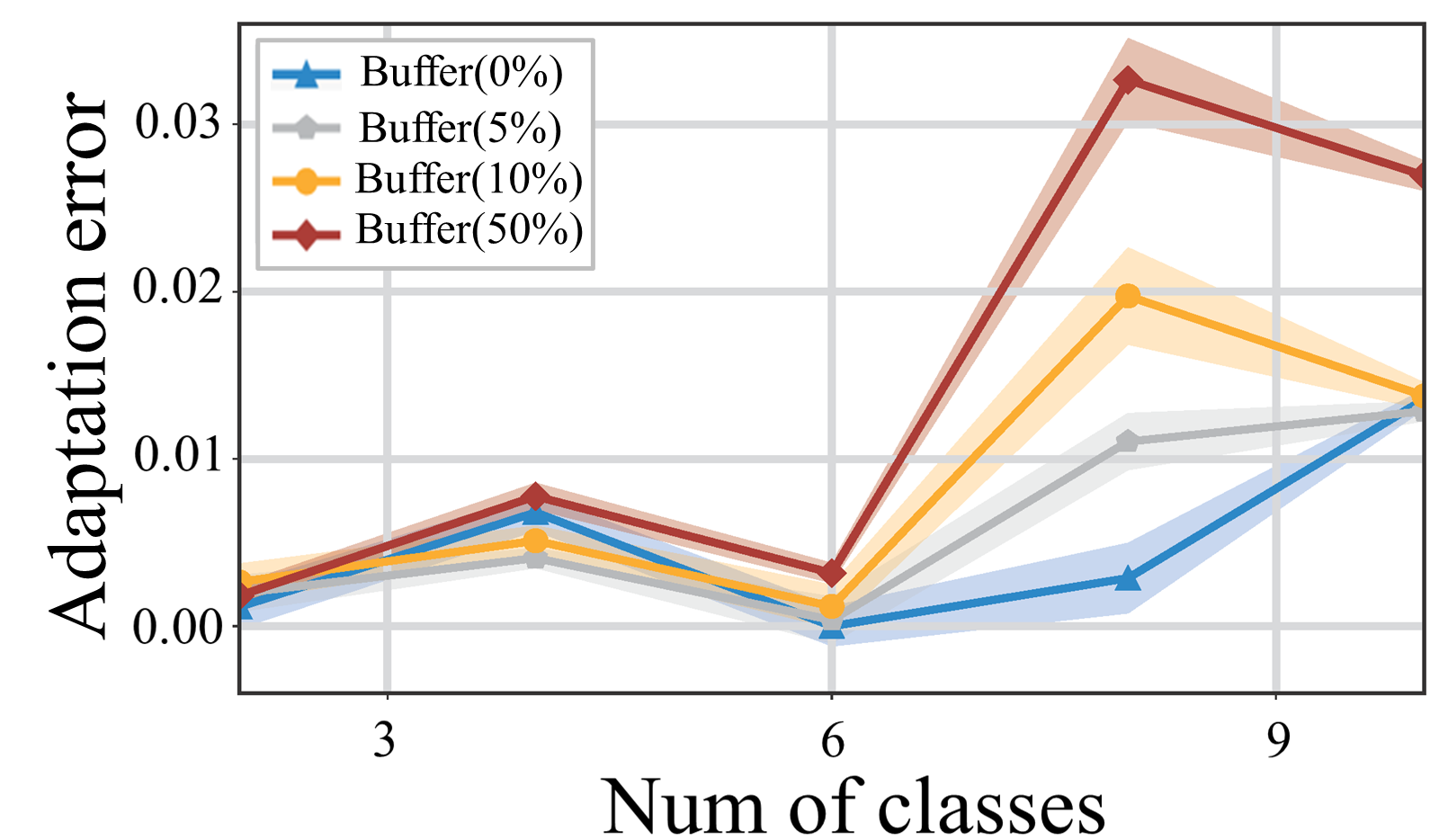}\\
            {\small (b) Adaptation error on MNIST}
            \vspace{4pt}
            
            \includegraphics[width=\linewidth]{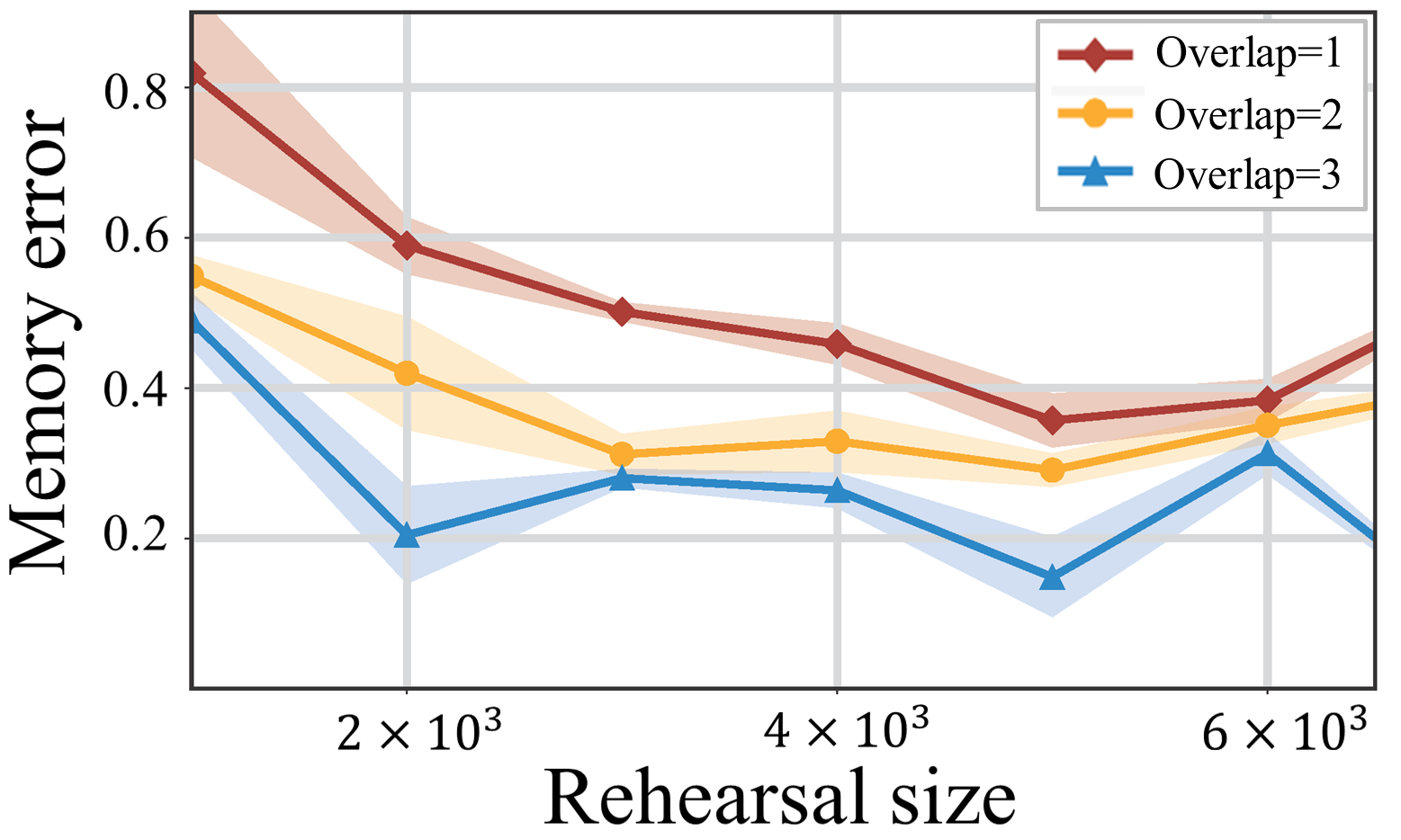}\\
            {\small (e) Memory error on CIFAR-10}
        \end{minipage}
        \hfill
        \begin{minipage}[t]{0.32\textwidth}
            \centering
            \includegraphics[width=\linewidth]{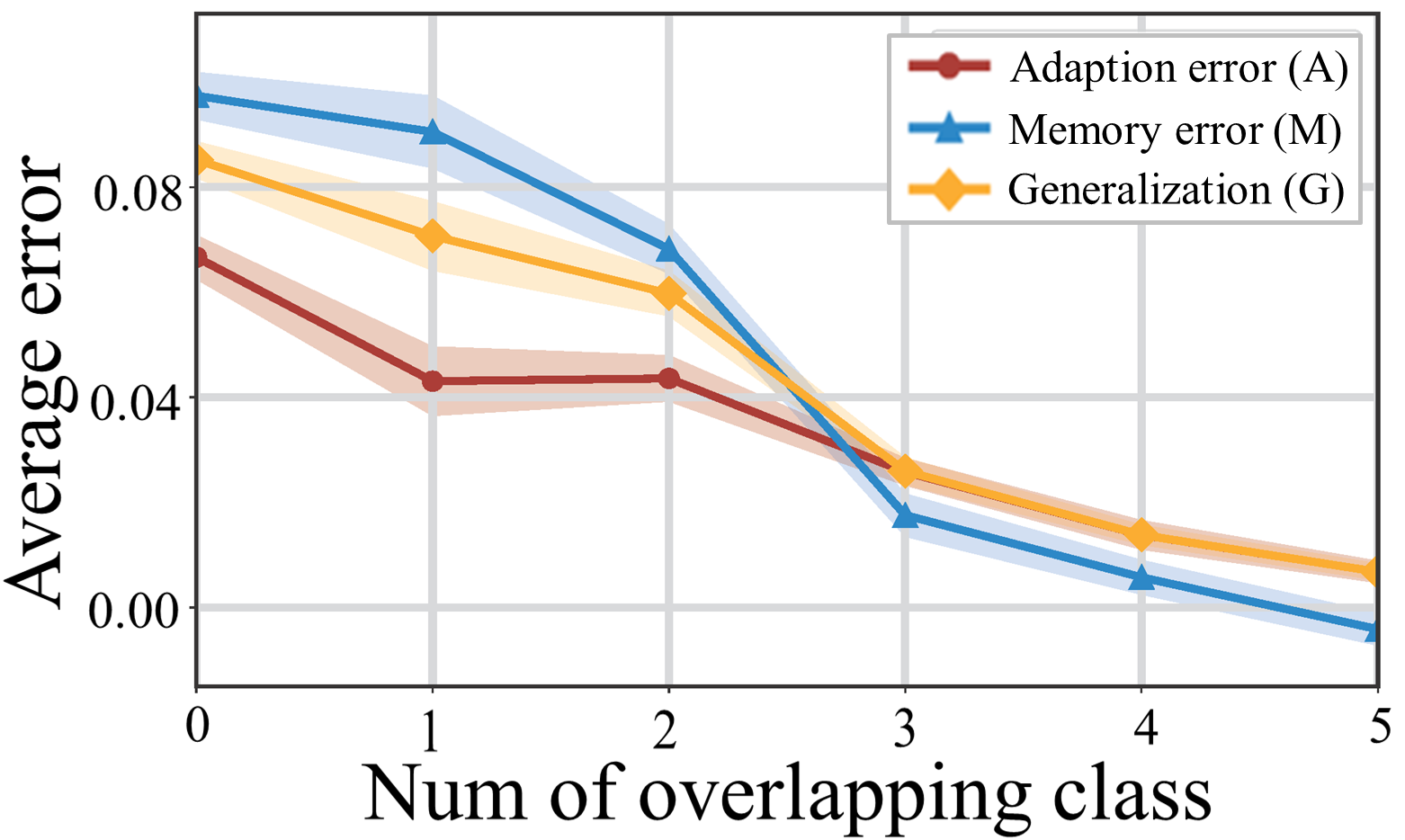}\\
            {\small (c) Average Error on MNIST}
           \vspace{4pt}
            
            \includegraphics[width=\linewidth]{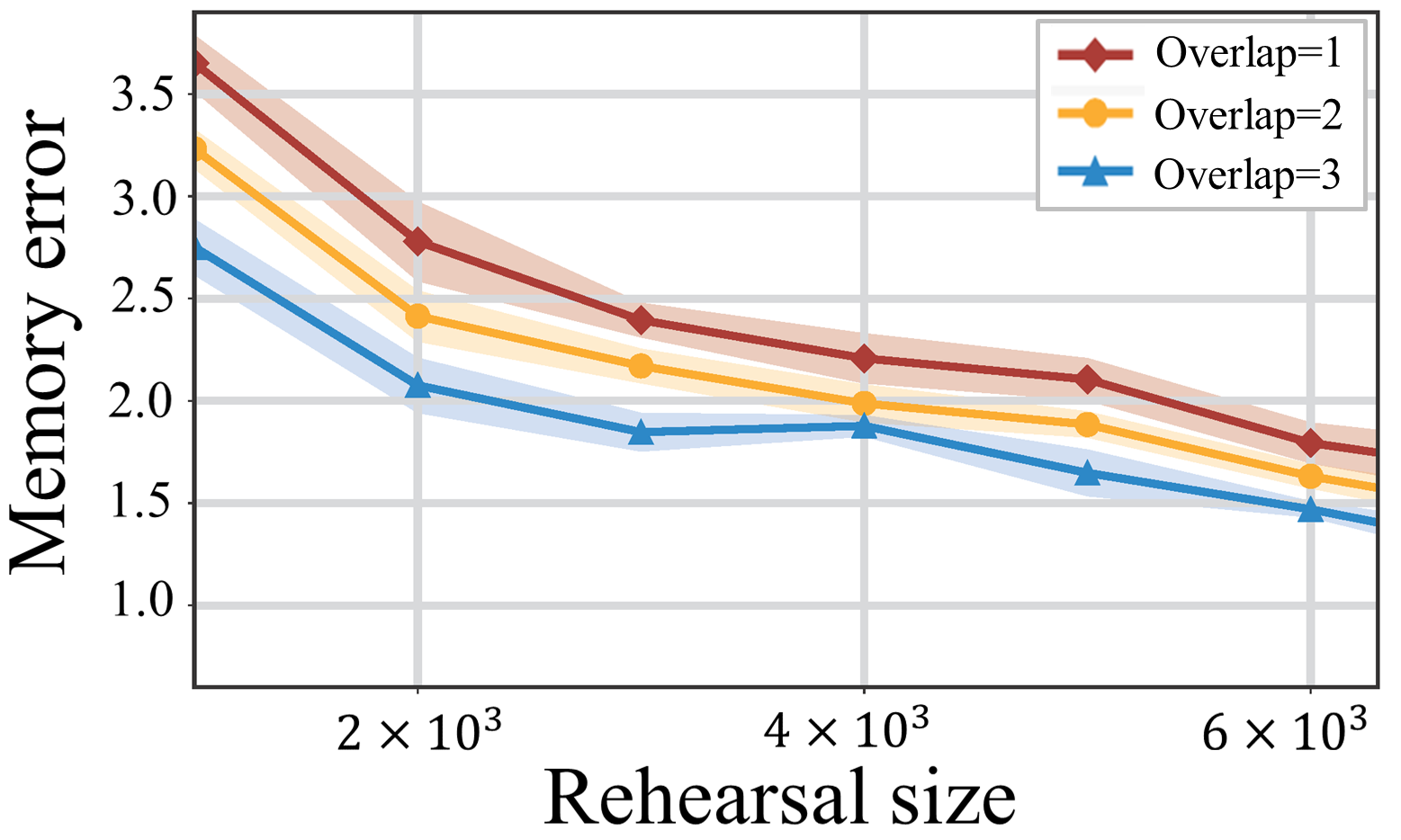}\\
            {\small (f) Memory error on CIFAR-100}
        \end{minipage}
        \caption{
        Impact of rehearsal size on adaptation, memory, and generalization errors
        in deep neural networks trained on MNIST, CIFAR-10, and CIFAR-100.
        Subfigures (a)-(b) show how accumulated classes affect adaptation error under different buffer sizes;
        (c)-(d) illustrate the impact of inter-task relatedness on generalization;
        (e)-(f) show how rehearsal size influences memory error under varying similarity levels.
        }
        \label{图片_DNN_threedatasets}
    \end{figure*}

    \begin{table*}[t]
    \centering
    \normalsize     
    \caption{Comparison of three types of model errors with traditional metrics as training classes increase on CIFAR-10.}
    \label{tab:error_metric_comparison}
    \begin{tabular}{lccccc}
    \toprule
    \textbf{Training Classes} & \textbf{2} & \textbf{4} & \textbf{6} & \textbf{8} & \textbf{10} \\
    \midrule
    Average Adaptation Error     & 0.127$\pm$0.004 & 0.398$\pm$0.003 & 0.279$\pm$0.001 & 0.105$\pm$0.003 & 0.159$\pm$0.002 \\
    Average Memory Error         & --              & 1.544$\pm$0.031 & 3.896$\pm$0.054 & 4.238$\pm$0.163 & 4.655$\pm$0.065 \\
    Average Generalization Error      & 0.127$\pm$0.004 & 1.035$\pm$0.013 & 2.865$\pm$0.038 & 3.406$\pm$0.121 & 3.938$\pm$0.053 \\
    Forgetting Ratio     & --              & 7.300$\pm$0.628 & 22.350$\pm$0.536 & 27.233$\pm$0.427 & 38.375$\pm$0.135 \\
    Average Accuracy     & 95.050$\pm$0.260 & 65.775$\pm$0.335 & 39.200$\pm$0.414 & 35.775$\pm$0.179 & 32.060$\pm$0.334 \\
    \bottomrule
    \end{tabular}
    \label{表格_误差指标对比}
    \end{table*}

    \begin{proposition}
    \label{proposition1}
    Assuming that Assumptions~\ref{assumption1} and~\ref{assumption2} hold. For $T=2$, the generalization error $\mathbb{E}[\mathcal G(\widehat{\boldsymbol{w}}_{T} )]$ increases with error on initial task when $\left \| \boldsymbol w_2^{*} -\boldsymbol w_1^{*}\right \|$ is low. Minimizing error requires small memory and adaptation errors while maintaining performance on the initial task.  
    \end{proposition}

    The proof is provided in Appendix D. As indicated by Proposition~\ref{proposition1}, better generalization requires excelling at current task while retaining knowledge from previous tasks. In addition, performance on initial task is also crucial, consistent with the empirical analyses by \cite{oracle4cvpr2022,1survey3tpami2024}. From the perspective of memorability, failing to learn initial task well can lead to error accumulation if knowledge retention is overemphasized.

\section{Empirical Validation on DNNs}
\label{section5_dnn experiments}

    Thus far, we have explored different aspects influencing the performance of rehearsal-based continual learning. To validate whether these theoretical insights under overparameterization extend to deep neural networks, we conduct experiments on standard real-world datasets. After training each task, the adaptation, memory, and generalization errors were systematically evaluated. The experiments were conducted on MNIST \cite{datamnist1989} , CIFAR-10 \cite{datacifar10and100in2009} , CIFAR-100 \cite{datacifar10and100in2009}, and Tiny-ImageNet \cite{datatinyimagenet2015}. All experiments were repeated at least three times, and average results are reported, with details provided in Appendix H.

    \begin{table}[h]
    \centering
    \normalsize    
    \caption{Comparison of current accuracy across continual learning benchmarks as rehearsal increases on CIFAR-100.}
    \label{tab:current_acc}
    \begin{tabular}{lccccc}
    \toprule
    \textbf{Methods} & \textbf{2000} & \textbf{4000} & \textbf{6000} & \textbf{8000} & \textbf{10000} \\
    \midrule
    EWC     & 86.20 & 84.60 & 83.17 & 81.91 & 80.11 \\
    LwF     & 73.35 & 73.86 & 72.92 & 73.90 & 74.76 \\
    iCaRL   & 84.59 & 82.98 & 81.59 & 80.89 & 80.19 \\
    DER     & 73.87 & 73.57 & 72.40 & 72.86 & 72.40 \\
    FOSTER  & 79.48 & 84.71 & 84.35 & 83.95 & 83.10 \\
    MEMO    & 68.10 & 66.78 & 66.01 & 65.56 & 65.16 \\
    CSReL   & 73.29 & 68.20 & 68.13 & 67.96 & 60.59 \\
    \bottomrule
    \end{tabular}
    \label{表格_综述当前准确率}
    \end{table}

    To examine how the effect of rehearsal on adaptation performance as the number of training classes increases, we partitioned MNIST and CIFAR-10 into 5 tasks, each task containing 2 classes. The CIFAR-100 and Tiny-ImageNet were divided into 10 and 20 tasks, respectively, with each task comprising 10 classes. Figures~\ref{图片_DNN_threedatasets}(a)–(b) and Figure~\ref{图片_DNN_tinyimagenet1eat} illustrate how the adaptation error varies with the number of training classes under different buffer sizes. As shown in the figures, the adaptation error consistently increases as the rehearsal scale grows (e.g., the red curves in the figures), indicating that rehearsal can hinder model adaptability. This observation consistent with the theoretical analysis.

    \graphicspath{{picture/}}
    \begin{figure}[h]
        \centering
        \includegraphics[width=\linewidth]{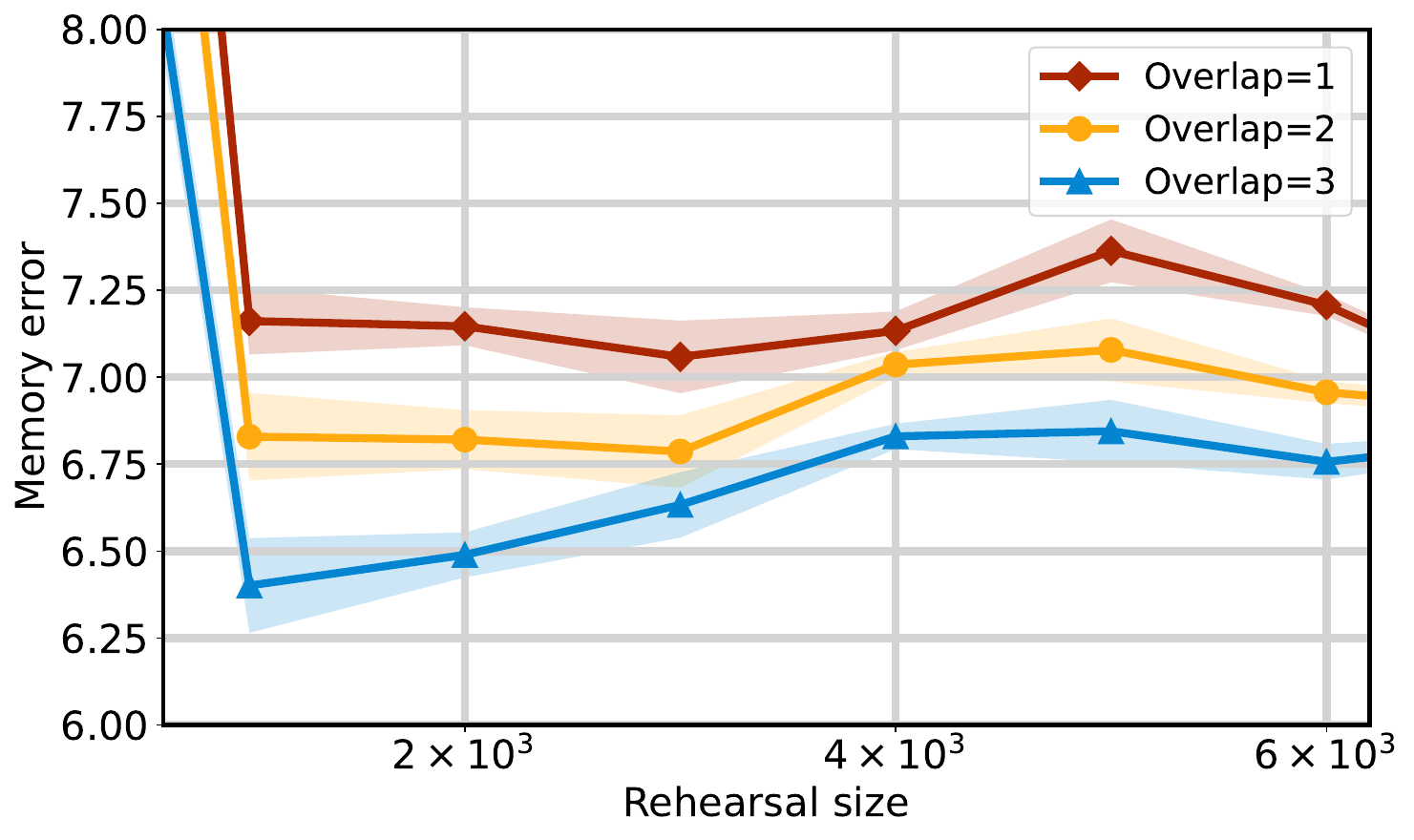}
        \caption{The memory error as rehearsal size increases across different levels of classes interrelation on Tiny-ImageNet.}
        \label{图片_DNN_tinyimagenet2emt}
    \end{figure}

    The impact of rehearsal size on memory errors is illustrated in Figures~\ref{图片_DNN_threedatasets}(e)–(f) and Figure~\ref{图片_DNN_tinyimagenet2emt}. Specifically, we split the MNIST and CIFAR-10 into 2 tasks, each comprising 5 classes. The partitioning schemes for CIFAR-100 and Tiny-ImageNet follows the previous settings. In the figures, the memory errors first decrease and then increase as the rehearsal size grows, with this effect being more pronounced when sharing category is two. These observations suggest that larger rehearsal do not necessarily lead to better memorability, and that further gains become marginal once rehearsal reaches a certain level.

    To examine the impact of inter-task relatedness on the three types of errors, we constructed datasets in which tasks share a varying proportion of classes. Results under additional evaluation metrics are reported in Appendix H. As shown in Figure~\ref{图片_DNN_threedatasets}(c)–(d), increasing task similarity consistently reduces adaptation error, memory error, and generalization error. Among these, memory error exhibits the most pronounced decrease (as indicated by the blue curve), suggesting that higher relatedness enhances knowledge retention.

    \begin{table}[H]
    \centering
    \normalsize     
    \caption{The average error comparing different sampling strategies and deeper network architectures on CIFAR-10.}
    \label{tab:combined_results}
    \begin{subtable}[t]{\linewidth}
    \centering
    \begin{tabular}{lcccc}
    \toprule
    \textbf{Methods} & 0\% & 5\% & 10\% & 50\% \\
    \midrule
    Random Sampling   & 0.196 & 0.212 & 0.257 & 0.382 \\
    Reservoir Sampling & 0.201 & 0.222 & 0.289 & 0.613 \\
    Herding Sampling  & 0.195 & 0.203 & 0.207 & 0.218 \\
    \specialrule{1pt}{0.1pt}{0.3em}
    \end{tabular}
    \caption{Adaptation error across sampling strategies.}
    \label{表格1_抽样策略}
    \end{subtable}
    \vspace{0.6em}
    \begin{subtable}[t]{\linewidth}
    \centering
    \begin{tabular}{lccc}
    \toprule
    \textbf{Arch.} & Adaptation & Memory & Generalization \\
    \midrule
    CNN      & 0.212 & 3.602 & 2.280 \\
    ResNet18 & 0.176 & 2.398 & 1.482 \\
    ResNet50 & 0.165 & 2.167 & 1.439 \\
    \specialrule{1pt}{0.1pt}{0.3em}
    \end{tabular}
    \caption{Average errors across network architectures.}
    \label{表格2_网络架构}
    \end{subtable}
    \end{table}  
    \vspace{-1em}

    Note that the model performance in deep neural network experiments is evaluated using the adaptation, memory, and generalization errors, to remain consistent with theoretical analysis, rather than accuracy or forgetting rate. Similar metrics are also used in research by \cite{kstatisticaltheory2024,khowcatastrophic2022}. Additionally, we also report accuracy and forgetting rate (Table~\ref{表格_误差指标对比}). As shown in the table, the average accuracy and generalization error reflect overall generalization, with accuracy gradually decreasing as training classes increase. And the forgetting rate and memory error indicate the ability to retain previous knowledge, both rising to demonstrate the phenomenon of forgetting.

    We extend the above analysis to existing continual learning baselines to further validate the above theoretical findings. The experimental setup follows~\cite{exper8_survey} to ensure fair comparisons. Specifically, we evaluate classical continual learning algorithms, including EWC~\cite{zhengzehua1EWC2017}, LwF~\cite{exper2_lwf}, iCaRL~\cite{exper3_icarl}, and DER~\cite{5forget1structure4cvpr2021}, alongside state-of-the-art approaches such as Foster~\cite{5forget2structure5eccv2022}, MEMO~\cite{exper6_memo}, and CSReL~\cite{exper7_CsRel}. Table~\ref{表格_综述当前准确率} reports the average current-task accuracy on CIFAR-100, while the results of forgetting rate are provided in Appendix H.

    As shown in Table~\ref{表格_综述当前准确率}, most algorithms experience some degree of accuracy decline when learning newly arrived tasks as the rehearsal increases. In particular, EWC and CSReL exhibit more pronounced drops, whereas DER and MEMO demonstrate relatively slower declines. Since CSReL constrains the model’s loss through an auxiliary network, which may interfere with task-specific training. These observations further highlight the limitations of rehearsal in facilitating adaptation to new tasks, emphasizing the importance of addressing this challenge in future algorithm designs.

    Building on the aforementioned results, we further examine the impact of different sampling strategies, network architectures, and longer task sequences. Table~\ref{表格1_抽样策略} compares model errors under random, herding, and reservoir sampling, showing that herding results in the lowest error. Table~\ref{表格2_网络架构} evaluates deeper network architectures, where deeper models consistently perform better, especially in memorability and generalization. Finally, we examine longer sequences on Tiny-ImageNet, with adaptation errors shown in Table~\ref{表格4_更多任务数量tinyimagenet}. In summary, adaptation errors increase as buffer size grows, across different sampling strategies, network depths, and task sequences. These results reinforce the previous analysis and highlight their consistency across various factors.

    In light of the above analysis, the impact of rehearsal scale on continual learning presents a double-edged sword. While rehearsal mechanism offers convenience, it can also introduce potential drawbacks. As demonstrated in this section, adding more replay data can hinder the model's adaptability to new tasks, and further replaying provides diminishing returns in terms of memorability. This conclusion is supported by experiments conducted across multiple datasets, varying sampling strategies, deeper network architectures, longer task sequences, and existing continual learning benchmarks, which collectively offer valuable insights for developing more advanced algorithms to overcome these challenges.

\section{Conclusion}
\label{section6_conclusion}

    In this work, we systematically investigate rehearsal scale in continual learning across three performance dimensions, modeling sequential tasks as Gaussian regression problems. By deriving explicit expressions for memory, adaptation, and generalization error with respect to rehearsal scale, our analysis reveals intriguing and counterintuitive insights regarding rehearsal scale: contrary to common assumptions, increasing the rehearsal scale harms the model’s adaptability, and even when mitigating forgetting, there exists a lower bound on memory error reduction. We validate these theoretical findings through numerical simulations and extend our analysis from theoretical models to deep neural networks. Experiments on real-world datasets further explore longer task sequences, deeper network architectures, and diverse sampling strategies. These results offer critical insights into both the benefits and inherent limitations of the rehearsal mechanism.

\bibliographystyle{IEEEtran}
\bibliography{cas-refs}

\begin{thebibliography}{10}
\providecommand{\url}[1]{#1}
\csname url@samestyle\endcsname
\providecommand{\newblock}{\relax}
\providecommand{\bibinfo}[2]{#2}
\providecommand{\BIBentrySTDinterwordspacing}{\spaceskip=0pt\relax}
\providecommand{\BIBentryALTinterwordstretchfactor}{4}
\providecommand{\BIBentryALTinterwordspacing}{\spaceskip=\fontdimen2\font plus
\BIBentryALTinterwordstretchfactor\fontdimen3\font minus \fontdimen4\font\relax}
\providecommand{\BIBforeignlanguage}[2]{{%
\expandafter\ifx\csname l@#1\endcsname\relax
\typeout{** WARNING: IEEEtran.bst: No hyphenation pattern has been}%
\typeout{** loaded for the language `#1'. Using the pattern for}%
\typeout{** the default language instead.}%
\else
\language=\csname l@#1\endcsname
\fi
#2}}
\providecommand{\BIBdecl}{\relax}
\BIBdecl

\bibitem{2continual1robotics1995}
S.~Thrun and T.~M. Mitchell, ``Lifelong robot learning,'' \emph{Robotics and autonomous systems}, vol.~15, no. 1-2, pp. 25--46, 1995.

\bibitem{2continual2aaai1986}
J.~C. Schlimmer and D.~Fisher, ``A case study of incremental concept induction,'' in \emph{Proceedings of the Fifth AAAI National Conference on Artificial Intelligence}, 1986, pp. 496--501.

\bibitem{3catastrophic1psychology1989}
M.~McCloskey and N.~J. Cohen, ``Catastrophic interference in connectionist networks: The sequential learning problem,'' in \emph{Psychology of learning and motivation}.\hskip 1em plus 0.5em minus 0.4em\relax Elsevier, 1989, vol.~24, pp. 109--165.

\bibitem{3catastrophic2arXiv2013}
I.~J. Goodfellow, M.~Mirza, D.~Xiao, A.~Courville, and Y.~Bengio, ``An empirical investigation of catastrophic forgetting in gradient-based neural networks,'' \emph{arXiv preprint arXiv:1312.6211}, 2013.

\bibitem{3catastrophic3iclr2020}
V.~V. Ramasesh, E.~Dyer, and M.~Raghu, ``Anatomy of catastrophic forgetting: Hidden representations and task semantics,'' in \emph{International Conference on Learning Representations}, 2021.

\bibitem{4tradeoff1neurosciences2005}
W.~C. Abraham and A.~Robins, ``Memory retention--the synaptic stability versus plasticity dilemma,'' \emph{Trends in neurosciences}, vol.~28, no.~2, pp. 73--78, 2005.

\bibitem{4tradeoff2zhengzehua2cvpr2022}
G.~Lin, H.~Chu, and H.~Lai, ``Towards better plasticity-stability trade-off in incremental learning: A simple linear connector,'' in \emph{Proceedings of the IEEE/CVF conference on computer vision and pattern recognition}, 2022, pp. 89--98.

\bibitem{4tradeoff3cvpr2023}
S.~Kim, L.~Noci, A.~Orvieto, and T.~Hofmann, ``Achieving a better stability-plasticity trade-off via auxiliary networks in continual learning,'' in \emph{Proceedings of the IEEE/CVF Conference on Computer Vision and Pattern Recognition}, 2023, pp. 11\,930--11\,939.

\bibitem{replay1GEM2017}
D.~Lopez-Paz and M.~Ranzato, ``Gradient episodic memory for continual learning,'' \emph{Advances in neural information processing systems}, vol.~30, 2017.

\bibitem{5forget1structure4cvpr2021}
S.~Yan, J.~Xie, and X.~He, ``Der: Dynamically expandable representation for class incremental learning,'' in \emph{Proceedings of the IEEE/CVF conference on computer vision and pattern recognition}, 2021, pp. 3014--3023.

\bibitem{5forget3zhengzehua5cvpr2023}
Z.~Sun, Y.~Mu, and G.~Hua, ``Regularizing second-order influences for continual learning,'' in \emph{Proceedings of the IEEE/CVF Conference on Computer Vision and Pattern Recognition}, 2023, pp. 20\,166--20\,175.

\bibitem{ktwoplayergame2021}
K.~Raghavan and P.~Balaprakash, ``Formalizing the generalization-forgetting trade-off in continual learning,'' \emph{Advances in Neural Information Processing Systems}, vol.~34, pp. 17\,284--17\,297, 2021.

\bibitem{6generate1cvpr2022}
C.~Simon, M.~Faraki, Y.-H. Tsai, X.~Yu, S.~Schulter, Y.~Suh, M.~Harandi, and M.~Chandraker, ``On generalizing beyond domains in cross-domain continual learning,'' in \emph{Proceedings of the IEEE/CVF conference on computer vision and pattern recognition}, 2022, pp. 9265--9274.

\bibitem{kforgetgenerate2023}
S.~Lin, P.~Ju, Y.~Liang, and N.~Shroff, ``Theory on forgetting and generalization of continual learning,'' in \emph{International Conference on Machine Learning}.\hskip 1em plus 0.5em minus 0.4em\relax PMLR, 2023, pp. 21\,078--21\,100.

\bibitem{7hippocampa1neuron2009}
T.~J. Davidson, F.~Kloosterman, and M.~A. Wilson, ``Hippocampal replay of extended experience,'' \emph{Neuron}, vol.~63, no.~4, pp. 497--507, 2009.

\bibitem{7hippocampa4science2025}
C.~S. Mallory, J.~Widloski, and D.~J. Foster, ``The time course and organization of hippocampal replay,'' \emph{Science}, vol. 387, no. 6733, pp. 541--548, 2025.

\bibitem{8rephippocampa1neurps2017}
H.~Shin, J.~K. Lee, J.~Kim, and J.~Kim, ``Continual learning with deep generative replay,'' \emph{Advances in neural information processing systems}, vol.~30, 2017.

\bibitem{replay2nature2020}
G.~M. Van~de Ven, H.~T. Siegelmann, and A.~S. Tolias, ``Brain-inspired replay for continual learning with artificial neural networks,'' \emph{Nature communications}, vol.~11, no.~1, p. 4069, 2020.

\bibitem{8rephippocampa3nature2025}
Q.~Shi, F.~Liu, H.~Li, G.~Li, L.~Shi, and R.~Zhao, ``Hybrid neural networks for continual learning inspired by corticohippocampal circuits,'' \emph{Nature Communications}, vol.~16, no.~1, p. 1272, 2025.

\bibitem{9improvereplay1ecv2018}
F.~M. Castro, M.~J. Mar{\'\i}n-Jim{\'e}nez, N.~Guil, C.~Schmid, and K.~Alahari, ``End-to-end incremental learning,'' in \emph{Proceedings of the European conference on computer vision (ECCV)}, 2018, pp. 233--248.

\bibitem{9improvereplay2replay3cvpr2022}
R.~Tiwari, K.~Killamsetty, R.~Iyer, and P.~Shenoy, ``Gcr: Gradient coreset based replay buffer selection for continual learning,'' in \emph{Proceedings of the IEEE/CVF Conference on Computer Vision and Pattern Recognition}, 2022, pp. 99--108.

\bibitem{9improvereplay3replay4icml2023}
R.~Gao and W.~Liu, ``Ddgr: Continual learning with deep diffusion-based generative replay,'' in \emph{International Conference on Machine Learning}.\hskip 1em plus 0.5em minus 0.4em\relax PMLR, 2023, pp. 10\,744--10\,763.

\bibitem{replay5neurps2024}
G.~Bellitto, F.~Proietto~Salanitri, M.~Pennisi, M.~Boschini, L.~Bonicelli, A.~Porrello, S.~Calderara, S.~Palazzo, and C.~Spampinato, ``Saliency-driven experience replay for continual learning,'' \emph{Advances in Neural Information Processing Systems}, vol.~37, pp. 103\,356--103\,383, 2024.

\bibitem{structure1PNN2016}
A.~A. Rusu, N.~C. Rabinowitz, G.~Desjardins, H.~Soyer, J.~Kirkpatrick, K.~Kavukcuoglu, R.~Pascanu, and R.~Hadsell, ``Progressive neural networks,'' \emph{arXiv preprint arXiv:1606.04671}, 2016.

\bibitem{structure2tnnls2022}
Q.~Gao, Z.~Luo, D.~Klabjan, and F.~Zhang, ``Efficient architecture search for continual learning,'' \emph{IEEE Transactions on Neural Networks and Learning Systems}, vol.~34, no.~11, pp. 8555--8565, 2022.

\bibitem{structure3cvpr2022}
A.~Douillard, A.~Ramé, G.~Couairon, and M.~Cord, ``Dytox: Transformers for continual learning with dynamic token expansion,'' in \emph{2022 IEEE/CVF Conference on Computer Vision and Pattern Recognition (CVPR)}, 2022, pp. 9275--9285.

\bibitem{structure6transip2024}
X.~Wang, Z.~Ji, Y.~Yu, Y.~Pang, and J.~Han, ``Model attention expansion for few-shot class-incremental learning,'' \emph{IEEE Transactions on Image Processing}, 2024.

\bibitem{pretrain2IJCV2025}
D.-W. Zhou, Z.-W. Cai, H.-J. Ye, D.-C. Zhan, and Z.~Liu, ``Revisiting class-incremental learning with pre-trained models: Generalizability and adaptivity are all you need,'' \emph{International Journal of Computer Vision}, vol. 133, no.~3, pp. 1012--1032, 2025.

\bibitem{pretrain4ICCV2023}
G.~Zhang, L.~Wang, G.~Kang, L.~Chen, and Y.~Wei, ``Slca: Slow learner with classifier alignment for continual learning on a pre-trained model,'' in \emph{Proceedings of the IEEE/CVF International Conference on Computer Vision}, 2023, pp. 19\,148--19\,158.

\bibitem{pretrain3CVPR2022}
Z.~Wang, Z.~Zhang, C.-Y. Lee, H.~Zhang, R.~Sun, X.~Ren, G.~Su, V.~Perot, J.~Dy, and T.~Pfister, ``Learning to prompt for continual learning,'' in \emph{Proceedings of the IEEE/CVF conference on computer vision and pattern recognition}, 2022, pp. 139--149.

\bibitem{pretrain1IJCAI2024}
D.-W. Zhou, H.-L. Sun, J.~Ning, H.-J. Ye, and D.-C. Zhan, ``Continual learning with pre-trained models: A survey,'' in \emph{IJCAI}, 2024.

\bibitem{zhengzehua1EWC2017}
J.~Kirkpatrick, R.~Pascanu, N.~Rabinowitz, J.~Veness, G.~Desjardins, A.~A. Rusu, K.~Milan, J.~Quan, T.~Ramalho, A.~Grabska-Barwinska \emph{et~al.}, ``Overcoming catastrophic forgetting in neural networks,'' \emph{Proceedings of the national academy of sciences}, vol. 114, no.~13, pp. 3521--3526, 2017.

\bibitem{zhengzehua3iclr2021}
A.~F. Aky{\"u}rek, E.~Aky{\"u}rek, D.~T. Wijaya, and J.~Andreas, ``Subspace regularizers for few-shot class incremental learning,'' \emph{International Conference on Learning Representations}, 2021.

\bibitem{zhengzehua4cvpr2023}
J.~Song, J.~Lee, I.~S. Kweon, and S.~Choi, ``Ecotta: Memory-efficient continual test-time adaptation via self-distilled regularization,'' in \emph{Proceedings of the IEEE/CVF Conference on Computer Vision and Pattern Recognition}, 2023, pp. 11\,920--11\,929.

\bibitem{1survey3tpami2024}
L.~Wang, X.~Zhang, H.~Su, and J.~Zhu, ``A comprehensive survey of continual learning: Theory, method and application,'' \emph{IEEE Transactions on Pattern Analysis and Machine Intelligence}, 2024.

\bibitem{kseparabledata2023}
I.~Evron, E.~Moroshko, G.~Buzaglo, M.~Khriesh, B.~Marjieh, N.~Srebro, and D.~Soudry, ``Continual learning in linear classification on separable data,'' in \emph{International Conference on Machine Learning}.\hskip 1em plus 0.5em minus 0.4em\relax PMLR, 2023, pp. 9440--9484.

\bibitem{kstatisticaltheory2024}
X.~Zhao, H.~Wang, W.~Huang, and W.~Lin, ``A statistical theory of regularization-based continual learning,'' \emph{arXiv preprint arXiv:2406.06213}, 2024.

\bibitem{ktheoreticalstudyood2022}
G.~Kim, C.~Xiao, T.~Konishi, Z.~Ke, and B.~Liu, ``A theoretical study on solving continual learning,'' \emph{Advances in neural information processing systems}, vol.~35, pp. 5065--5079, 2022.

\bibitem{kidealcontinual2023}
L.~Peng, P.~Giampouras, and R.~Vidal, ``The ideal continual learner: An agent that never forgets,'' in \emph{International Conference on Machine Learning}.\hskip 1em plus 0.5em minus 0.4em\relax PMLR, 2023, pp. 27\,585--27\,610.

\bibitem{khowcatastrophic2022}
I.~Evron, E.~Moroshko, R.~Ward, N.~Srebro, and D.~Soudry, ``How catastrophic can catastrophic forgetting be in linear regression?'' in \emph{Conference on Learning Theory}.\hskip 1em plus 0.5em minus 0.4em\relax PMLR, 2022, pp. 4028--4079.

\bibitem{kfixdesign2023}
H.~Li, J.~Wu, and V.~Braverman, ``Fixed design analysis of regularization-based continual learning,'' in \emph{Conference on Lifelong Learning Agents}.\hskip 1em plus 0.5em minus 0.4em\relax PMLR, 2023, pp. 513--533.

\bibitem{compare2arxiv2025}
J.~Deng, Q.~Wu, P.~Ju, S.~Lin, Y.~Liang, and N.~Shroff, ``Unlocking the power of rehearsal in continual learning: A theoretical perspective,'' \emph{arXiv preprint arXiv:2506.00205}, 2025.

\bibitem{compare3arxiv2025}
G.~Zheng, P.~Wang, and L.~Shen, ``Towards understanding memory buffer based continual learning,'' \emph{Neural Networks}, 2024.

\bibitem{compare4arxiv2024}
M.~Ding, K.~Ji, D.~Wang, and J.~Xu, ``Understanding forgetting in continual learning with linear regression,'' \emph{arXiv preprint arXiv:2405.17583}, 2024.

\bibitem{compare5icais2023}
D.~Goldfarb and P.~Hand, ``Analysis of catastrophic forgetting for random orthogonal transformation tasks in the overparameterized regime,'' in \emph{International Conference on Artificial Intelligence and Statistics}.\hskip 1em plus 0.5em minus 0.4em\relax PMLR, 2023, pp. 2975--2993.

\bibitem{compare1mubiaoarxiv2024}
A.~Banayeeanzade, M.~Soltanolkotabi, and M.~Rostami, ``Theoretical insights into overparameterized models in multi-task and replay-based continual learning,'' \emph{arXiv preprint arXiv:2408.16939}, 2024.

\bibitem{normal1NIPS2023}
A.~Ravent{\'o}s, M.~Paul, F.~Chen, and S.~Ganguli, ``Pretraining task diversity and the emergence of non-bayesian in-context learning for regression,'' \emph{Advances in neural information processing systems}, vol.~36, pp. 14\,228--14\,246, 2023.

\bibitem{normal2ICLR2025}
H.~Li, S.~Lin, L.~Duan, Y.~Liang, and N.~Shroff, ``Theory on mixture-of-experts in continual learning,'' in \emph{The Thirteenth International Conference on Learning Representations}, 2025.

\bibitem{linear1JMLR2023}
W.~Ji, Z.~Deng, R.~Nakada, J.~Zou, and L.~Zhang, ``The power of contrast for feature learning: A theoretical analysis,'' \emph{Journal of Machine Learning Research}, vol.~24, no. 330, pp. 1--78, 2023.

\bibitem{1survey1nn2019}
G.~I. Parisi, R.~Kemker, J.~L. Part, C.~Kanan, and S.~Wermter, ``Continual lifelong learning with neural networks: A review,'' \emph{Neural networks}, vol. 113, pp. 54--71, 2019.

\bibitem{1survey2tpami2021}
M.~De~Lange, R.~Aljundi, M.~Masana, S.~Parisot, X.~Jia, A.~Leonardis, G.~Slabaugh, and T.~Tuytelaars, ``A continual learning survey: Defying forgetting in classification tasks,'' \emph{IEEE transactions on pattern analysis and machine intelligence}, vol.~44, no.~7, pp. 3366--3385, 2021.

\bibitem{estimatereplay1neurps2018}
D.~Rolnick, A.~Ahuja, J.~Schwarz, T.~P. Lillicrap, and G.~Wayne, ``Experience replay for continual learning,'' in \emph{Neural Information Processing Systems}, 2018.

\bibitem{estimatereplay2arxiv2022}
L.~Wang, X.~Zhang, K.~Yang, L.~L. Yu, C.~Li, L.~Hong, S.~Zhang, Z.~Li, Y.~Zhong, and J.~Zhu, ``Memory replay with data compression for continual learning,'' \emph{ArXiv}, vol. abs/2202.06592, 2022.

\bibitem{estimatereplay3arxiv2023}
K.~Jeeveswaran, P.~Bhat, B.~Zonooz, and E.~Arani, ``Birt: Bio-inspired replay in vision transformers for continual learning,'' \emph{ArXiv}, vol. abs/2305.04769, 2023.

\bibitem{oracle4cvpr2022}
Y.~Shi, K.~Zhou, J.~Liang, Z.~Jiang, J.~Feng, P.~H. Torr, S.~Bai, and V.~Y. Tan, ``Mimicking the oracle: An initial phase decorrelation approach for class incremental learning,'' in \emph{Proceedings of the IEEE/CVF conference on computer vision and pattern recognition}, 2022, pp. 16\,722--16\,731.

\bibitem{datamnist1989}
Y.~LeCun, B.~Boser, J.~Denker, D.~Henderson, R.~Howard, W.~Hubbard, and L.~Jackel, ``Handwritten digit recognition with a back-propagation network,'' \emph{Advances in neural information processing systems}, vol.~2, 1989.

\bibitem{datacifar10and100in2009}
A.~Krizhevsky, G.~Hinton \emph{et~al.}, ``Learning multiple layers of features from tiny images,'' 2009.

\bibitem{datatinyimagenet2015}
Y.~Le and X.~Yang, ``Tiny imagenet visual recognition challenge,'' \emph{CS 231N}, vol.~7, no.~7, p.~3, 2015.

\bibitem{exper8_survey}
D.-W. Zhou, Q.-W. Wang, Z.-H. Qi, H.-J. Ye, D.-C. Zhan, and Z.~Liu, ``Class-incremental learning: A survey,'' \emph{IEEE Transactions on Pattern Analysis and Machine Intelligence}, 2024.

\bibitem{exper2_lwf}
Z.~Li and D.~Hoiem, ``Learning without forgetting,'' \emph{IEEE transactions on pattern analysis and machine intelligence}, vol.~40, no.~12, pp. 2935--2947, 2017.

\bibitem{exper3_icarl}
S.-A. Rebuffi, A.~Kolesnikov, G.~Sperl, and C.~H. Lampert, ``icarl: Incremental classifier and representation learning,'' in \emph{Proceedings of the IEEE conference on Computer Vision and Pattern Recognition}, 2017, pp. 2001--2010.

\bibitem{5forget2structure5eccv2022}
F.-Y. Wang, D.-W. Zhou, H.-J. Ye, and D.-C. Zhan, ``Foster: Feature boosting and compression for class-incremental learning,'' in \emph{European conference on computer vision}.\hskip 1em plus 0.5em minus 0.4em\relax Springer, 2022, pp. 398--414.

\bibitem{exper6_memo}
D.-W. Zhou, Q.-W. Wang, H.-J. Ye, and D.-C. Zhan, ``A model or 603 exemplars: Towards memory-efficient class-incremental learning,'' \emph{arXiv preprint arXiv:2205.13218}, 2022.

\bibitem{exper7_CsRel}
R.~Tong, Y.~Liu, J.~Q. Shi, and D.~Gong, ``Coreset selection via reducible loss in continual learning,'' in \emph{The Thirteenth International Conference on Learning Representations}, 2025.

\end{thebibliography}

\end{document}